\documentclass[]{jingdong}

\usepackage{amsmath,amsfonts,bm}

\def\eqref#1{equation~\ref{#1}}

\def\1{\bm{1}}

\DeclareMathAlphabet{\mathsfit}{\encodingdefault}{\sfdefault}{m}{sl}
\SetMathAlphabet{\mathsfit}{bold}{\encodingdefault}{\sfdefault}{bx}{n}

\usepackage{url}
\usepackage{booktabs}
\usepackage{array}
\usepackage{graphicx}
\usepackage{float}
\usepackage{flafter}
\usepackage{placeins}
\usepackage{colortbl}
\usepackage{microtype}
\usepackage{enumitem}
\usepackage{multicol}
\usepackage{fontawesome5}  %
\usepackage{pgfplots}
\pgfplotsset{compat=1.18}
\usetikzlibrary{patterns}

\definecolor{watermelonlight}{RGB}{252,222,224}
\definecolor{watermelonpale}{RGB}{255,242,243}
\definecolor{watermelondark}{RGB}{176,45,58}

\title{MIRA: A Multimodal Illustrated Deep Research Agent for Controllable Reports}

\author[1,3,*]{Hongrui Wu}
\author[2,*]{Kaiwen Tuo}
\author[3,*]{Congcong Wang}
\author[3]{Gen Li}
\author[4]{Shuai Dong}
\author[5]{Xinlei Yu}
\author[3]{Yuqi Xu}
\author[6]{Haowen Hou}
\author[3]{Zelin Peng}
\author[3,\dagger]{Jiaqi Wang}

\affiliation[1]{Stanford University}
\affiliation[2]{The Hong Kong University of Science and Technology}
\affiliation[3]{JD.com}
\affiliation[4]{Fudan University}
\affiliation[5]{The Chinese University of Hong Kong}
\affiliation[6]{Shanghai Jiao Tong University}

\contribution[*]{Equal contribution}
\contribution[\dagger]{Corresponding author.}
\DeclareRobustCommand{\githublogo}{\faGithub}
\DeclareRobustCommand{\hflogo}{\raisebox{-0.15em}{\includegraphics[height=0.95em]{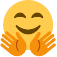}}}
\DeclareRobustCommand{\linkurl}[1]{{\checkdatafont\ttfamily #1}}
\DeclareRobustCommand{\linklabel}[1]{{\checkdatafont\sffamily\bfseries #1:}}
\checkdata[\githublogo\ Project Page]{\href{https://mira-mmdeepresearch.github.io/}{\linkurl{mira-mmdeepresearch.github.io}}%
  \hspace{1em}\linklabel{\hflogo\ Hugging Face} \href{https://huggingface.co/MIRA-MMDeepResearch}{\linkurl{huggingface.co/MIRA-MMDeepResearch}}}

\abstract{\setlength{\parskip}{2pt}
Deep research agents are increasingly capable of searching the open web and
synthesizing evidence, yet they remain poorly suited to \textbf{controllable
report generation in multimodal deep research}. This task requires agents to
handle interleaved textual and visual inputs, gather multimodal evidence through
multi-step web search, and deliver comprehensive illustrated reports under
user-specified requirements. Existing systems face two limitations. During
reasoning, they often underuse images discovered along the search trajectory;
during generation, they tend to produce fixed-form answers with limited control
over report granularity, multimodal composition, and evidence grounding. We
introduce \textbf{MIRA}, a \textbf{M}ultimodal \textbf{I}llustrated Deep
\textbf{R}esearch \textbf{A}gent for controllable reports. MIRA integrates
textual and visual evidence throughout research and writing. At its core,
MIRA-Harness translates user requirements into generation contracts and
connects four stages: evidence collection, report submission, feedback-driven
revision, and final delivery. Building on this harness, MIRA-Data constructs
high-quality multi-hop multimodal research trajectories, and we post-train
foundation models to operate within the workflow. Using Qwen3.6-35B-A3B, MIRA
achieves \textbf{state-of-the-art performance} among open agents on 7
representative, established multimodal-search and deep-research-agent
benchmarks, while producing user-aligned illustrated reports with traceable
textual and visual evidence.
}

\newcommand{\miraTeaserBeforeIntroduction}{}

\begin{document}

\maketitle

\vspace*{-3mm}
\begin{minipage}{\textwidth}
\centering
\includegraphics[width=\textwidth]{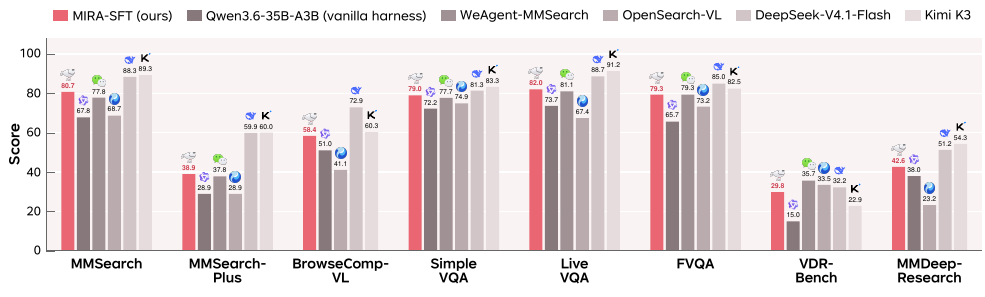}
\captionof{figure}{\textbf{Benchmark performance of MIRA.} MIRA-SFT versus open agents and frontier models (Table~\ref{tab:cross-work}).}
\label{fig:results-overview}
\end{minipage}

\clearpage
\ifdefined\miraTeaserBeforeIntroduction
\begin{figure}[H]
\else
\begin{figure}[t]
\fi
\centering
\includegraphics[width=\textwidth]{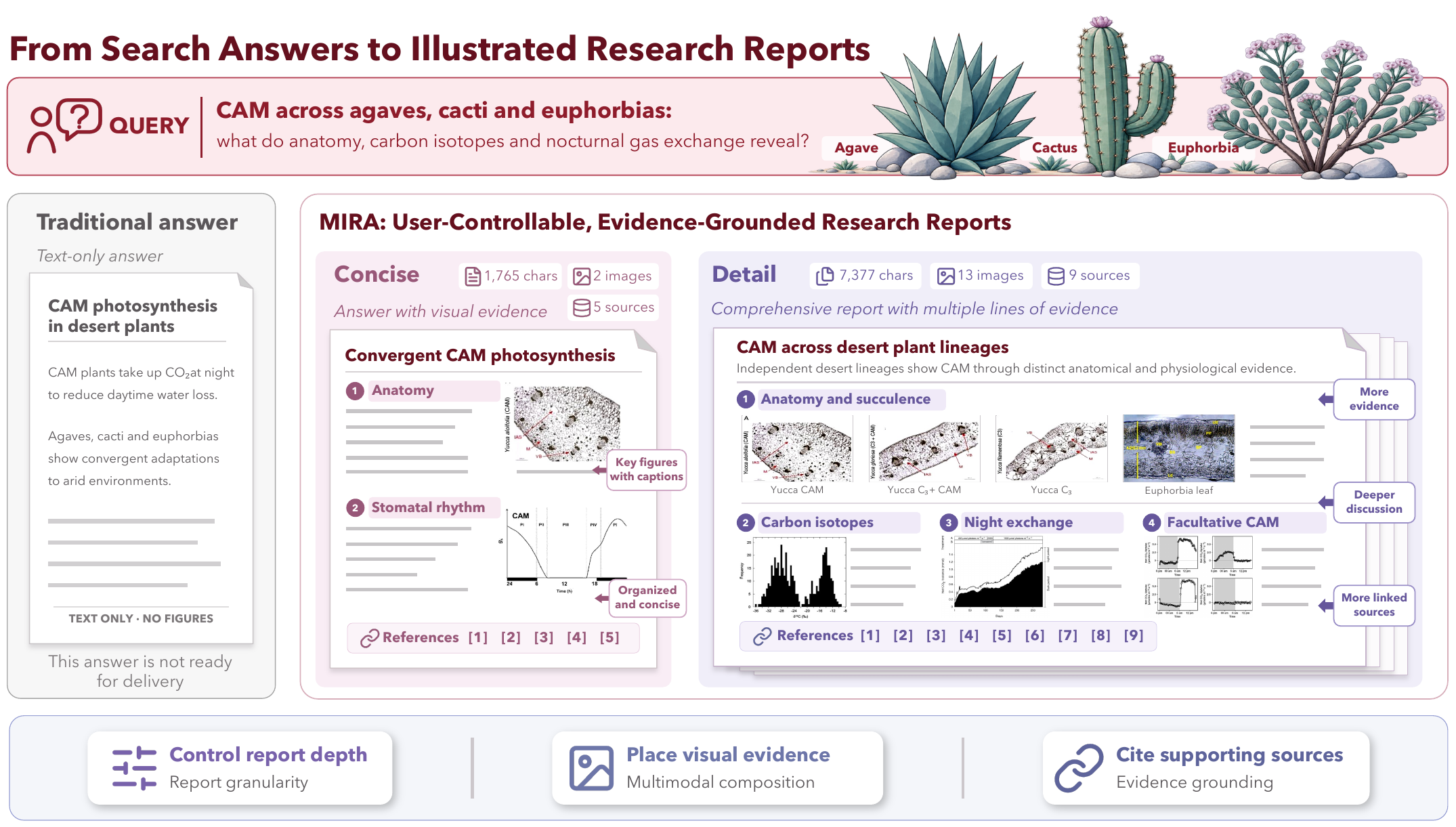}
\caption{\textbf{From search answers to controllable illustrated reports.}
For one request, MIRA controls report granularity, multimodal composition, and
evidence grounding across Concise and Detail deliverables.}
\label{fig:task-overview}
\end{figure}

\section{Introduction}
\label{sec:introduction}

Multimodal large language models (MLLMs) acquire extensive visual and textual
knowledge from static training corpora~\citep{alayrac2022flamingo,li2023blip,liu2023visual,Bai2023QwenVLAV,bai2025qwen3}. External tools extend this fixed
knowledge boundary, enabling models to retrieve evidence and address
real-world information needs~\citep{yao2023react,yang2023mmreact,jin2025searchr1,chen2026opensearchvl,weagent_manuscript}.
Text-based deep research systems such as STORM use multi-step retrieval to
write long-form articles~\citep{shao2024storm}. WebThinker trains an agent to
interleave web search and report drafting~\citep{li2025webthinker}.
Multimodal deep research extends this setting to textual and visual
evidence~\citep{huang2026mmdeepresearchbench}.
However, tool access alone does not ensure an illustrated report that simultaneously
provides accurate and sufficiently complete information~\citep{li2025reportbench},
integrates textual and visual evidence throughout research and
writing~\citep{pham-etal-2026-analystbench,liu-etal-2026-evireport}, and follows the user's
preferred style.

Motivated by real-world reporting needs, we study \textbf{controllable
report generation}: producing illustrated reports that respond to user
requirements in multimodal deep research. Given interleaved textual and visual
inputs alongside explicit delivery requirements, an agent must gather evidence
from the open web and satisfy constraints on three attributes: \textbf{report granularity}, the
scope, depth, and level of detail; \textbf{multimodal composition}, the
selection, balance, and organization of text and visuals; and \textbf{evidence
grounding}, the required evidential support and explicit links from textual
claims to source documents~\citep{gao-etal-2023-enabling} and from report
visuals to their original source figures~\citep{shi2026vidr}.

\ifdefined\miraTeaserBeforeIntroduction\else

\fi

Recent work addresses important components of this task. Vision Search
Assistant~\citep{zhang2024visionsearchassistant} uses external search to answer
questions about visual inputs. MMSearch-R1~\citep{wu2025mmsearchr1}
and Vision-DeepResearch~\citep{huang2026visiondeepresearch} integrate textual
and visual search into multi-step reasoning to answer visual questions that
require external knowledge. Moving toward report generation~\citep{ye-etal-2026-deep}, Multimodal
DeepResearcher~\citep{yang2025multimodaldeepresearcher} explores interleaved
text and chart generation, ViDR~\citep{shi2026vidr} grounds reports in source
figures, and Ptah~\citep{zhang2026ptah} coordinates illustrated report
generation and verification through a multi-agent harness. Together, these
advances provide a foundation for studying how different reporting needs can
be supported within a unified research workflow.

Nevertheless, jointly controlling report granularity, multimodal composition,
and evidence grounding remains challenging. Even for the same research
question, a concise brief and a detailed report may require different evidence
coverage, visual choices, and levels of exposition. These requirements should
therefore shape the research process, rather than serve only as formatting
instructions during final generation. This raises a central question:
\textbf{How should the user's intended deliverable guide evidence collection,
report organization, and the decision to conclude research?}

To address this question, we present \textbf{MIRA}, a
\textbf{M}ultimodal \textbf{I}llustrated Deep \textbf{R}esearch
\textbf{A}gent for controllable reports. At its core,
MIRA-Harness connects evidence collection, illustrated writing, and
delivery checks in an iterative loop. The harness represents user requirements
as \textbf{generation contracts}, which specify explicit conditions for accepting a
report, and supports \textbf{Concise} and \textbf{Detail} modes for different
reporting needs (Figure~\ref{fig:task-overview}). Both modes specify
requirements across all three attributes while preserving a shared requirement
for source traceability. During research, the agent inspects and records
sources and visual materials. After drafting, it submits the report for
checking, uses unmet conditions as revision targets, and revises the content or
gathers additional evidence before resubmitting within the allocated budget. A
report is accepted as a completed deliverable only after passing the
implemented contract checks.

To teach models this complete workflow, we develop MIRA-Data, a
framework for constructing report-oriented training trajectories that connect
evidence gathering, writing, submission, and feedback-driven revision. Using
Qwen3.6-35B-A3B as the backbone, we compare the unadapted model (Base) with
its supervised fine-tuned counterpart (SFT). Our evaluation examines both
report delivery and multimodal search,
assessing how adaptation affects report quality, adherence to user
requirements, and the ability to acquire evidence.

\begin{samepage}
Our contributions are threefold:
\begin{itemize}[noitemsep,topsep=0pt,parsep=0pt,partopsep=0pt, leftmargin=*]
    \item \textbf{Task formulation.} We identify the gap between multimodal
    deep research and real-world report delivery, and formulate controllable
    report generation: producing illustrated, evidence-grounded reports under
    user-specified requirements for report granularity, multimodal composition,
    and evidence grounding.
    \item \textbf{Harness and training data.} We introduce MIRA-Harness, which
    unifies user requirements, multimodal evidence collection, illustrated
    writing, and feedback-driven revision through generation contracts. We
    further develop MIRA-Data to construct training trajectories aligned with
    this workflow, covering evidence gathering, writing, submission checks,
    and revision.
    \item \textbf{Empirical evaluation.} We train and evaluate MIRA using
    Qwen3.6-35B-A3B, comparing Base and SFT variants across report delivery
    and multimodal search. Across eight multimodal search, knowledge-intensive VQA, 
    and illustrated deep-research benchmarks, MIRA-SFT achieves state-of-the-art
    performance among the compared open agents on seven of the eight benchmarks.
    Beyond benchmark performance, MIRA can generate comprehensive illustrated reports that adapt to user-specified
     requirements for report granularity, multimodal composition, and evidence grounding,
     while integrating visual evidence and traceable sources into the final deliverable.
\end{itemize}
\end{samepage}

\section{Related Work}
\label{sec:related-work}

\subsection{Deep Research and Report Generation}

Deep research extends tool-augmented language models from single-step retrieval to
long-horizon information seeking, where an agent iteratively searches, inspects external
evidence, and synthesizes an answer. Early agentic formulations such as ReAct
\citep{yao2023react} established the general paradigm of interleaving reasoning with
environmental actions, while systems such as STORM \citep{shao2024storm} demonstrated
that multi-stage retrieval and organization can support substantially longer and more
structured textual outputs. More recent work has shifted attention toward training search agents
directly, including scalable long-horizon search and synthetic supervision
\citep{chu2026redsearcher,xie2026quest}, showing that effective research behavior can be
learned, not merely prompted.

\subsection{Multimodal Search Agents}

Multimodal research agents further expand this setting by allowing visual observations to
participate in the search process. Benchmarks such as MMSearch, MMSearch-Plus, and
MM-BrowseComp \citep{jiang2024mmsearch,tao2025mmsearchplus,li2025mmbrowsecomp}
expose the need to retrieve and reason
over both textual and visual evidence, while MMSearch-R1 \citep{wu2025mmsearchr1},
WebWatcher \citep{webwatcher}, OpenSearch-VL \citep{chen2026opensearchvl},
Vision-DeepResearch \citep{huang2026visiondeepresearch}, and WeAgent-MMSearch
\citep{weagent_manuscript} develop increasingly capable multimodal browsing and search
policies. These works substantially improve how agents acquire external evidence,
particularly for knowledge-intensive visual questions.

\subsection{Agent Harnesses and Tool-Use Supervision}

The behavior of a research agent is determined not only by its underlying model, but also
by the execution environment that exposes tools, maintains interaction state, and
mediates long-horizon reasoning. Existing search systems therefore use tool environments
and agent harnesses to coordinate retrieval and multimodal inspection
\citep{chen2026opensearchvl,weagent_manuscript}. The WeAgent harness additionally exposes
webpage extraction and code execution~\citep{weagent_manuscript}. In parallel, recent work on search-agent
training constructs trajectories of tool calls and intermediate observations so that
models can internalize effective interaction patterns
\citep{chu2026redsearcher,xie2026quest}. This line of work shows that runtime structure 
and training supervision are complementary. A capable
backbone may still fail to search effectively when the interaction protocol is poorly
specified, while training on trajectories mismatched to the deployment environment can
teach behaviors that are difficult to realize at inference time.

\subsection{Controllable and Evidence-Grounded Report Generation}

A separate line of research moves beyond short-form question answering toward long-form,
evidence-backed synthesis. FActScore \citep{min-etal-2023-factscore} evaluates the
factual support of claims in long-form text. Citation-aware generation studies how
model outputs can be explicitly linked to retrieved sources
\citep{gao-etal-2023-enabling}, while systems such as
EviReport \citep{liu-etal-2026-evireport} organize evidence throughout long-form report
generation. Deep-Reporter \citep{ye-etal-2026-deep} further studies grounded multimodal
long-form generation. ReportBench \citep{li2025reportbench} evaluates the content
quality of research reports for academic survey tasks, while AnalystBench
\citep{pham-etal-2026-analystbench} evaluates professional report generation in settings
where completeness and multimodal presentation
matter in addition to factual correctness. These efforts highlight that high-quality
research outputs require more than finding a correct answer: evidence must be selected,
organized, and preserved across a substantially longer generation process.

Recent multimodal deep-research systems make this connection even more explicit.
Multimodal DeepResearcher \citep{yang2025multimodaldeepresearcher} studies interleaved
text-chart report generation, ViDR \citep{shi2026vidr} grounds generated reports in
source visual evidence, and Ptah \citep{zhang2026ptah} introduces a multi-agent harness
for interleaved report generation and verification. MMDeepResearch-Bench
\citep{huang2026mmdeepresearchbench} correspondingly evaluates multimodal agents on
long-form research synthesis. Collectively, these works establish the importance of
illustration, provenance, and verification for multimodal reports. Accordingly,
MIRA targets report delivery under explicit user requirements.

\section{Task Formulation}
\label{sec:task-formulation}
\raggedbottom

\subsection{Multimodal Deep Research}
\label{sec:background}

In tool-using multimodal deep research, an \textbf{agent harness} mediates
long-horizon interactions between a model and its environment.
This execution layer routes actions, returns observations, maintains state,
and enforces budgets~\citep{weagent_manuscript,chen2026opensearchvl}.
Given a possibly multimodal request \(q\) and model-visible history \(h_t\),
the policy \(\pi_\theta\), parameterized by \(\theta\), selects a tool action
\(a_t\) at turn \(t\), and the harness \(\mathcal H\) returns an observation
\(o_t\). Let \(\oplus\) append an action and observation pair to the history. The
interaction loop~\citep{yao2023react} is
\begin{equation}
\label{eq:agent-loop}
a_t\sim\pi_\theta(\cdot\mid q,h_t),\qquad
o_t=\mathcal H(a_t,h_t),\qquad
h_{t+1}=h_t\oplus(a_t,o_t).
\end{equation}
Here, \(o_t\) may contain text, image pixels, or tool errors. Over \(T\) turns,
the complete interaction forms a trajectory \(\tau\); the shorthand
\(\tau=(a_t,o_t)_{t=1}^{T}\) omits intervening model text and internal harness
state.
The harness thus structures the agent's evidence-gathering loop,
rather than merely exposing a list of tools.

\subsection{Controllable Illustrated Report Generation}

Building on this loop, controllable report generation in multimodal deep
research requires an illustrated, evidence-grounded report under user-specified
requirements, so delivery requirements must shape evidence gathering and
revision. MIRA-Harness incorporates these requirements into runtime
interaction, while MIRA-Data supplies trajectories for the same workflow.

\begin{figure}[t]
\centering
\includegraphics[width=\linewidth]{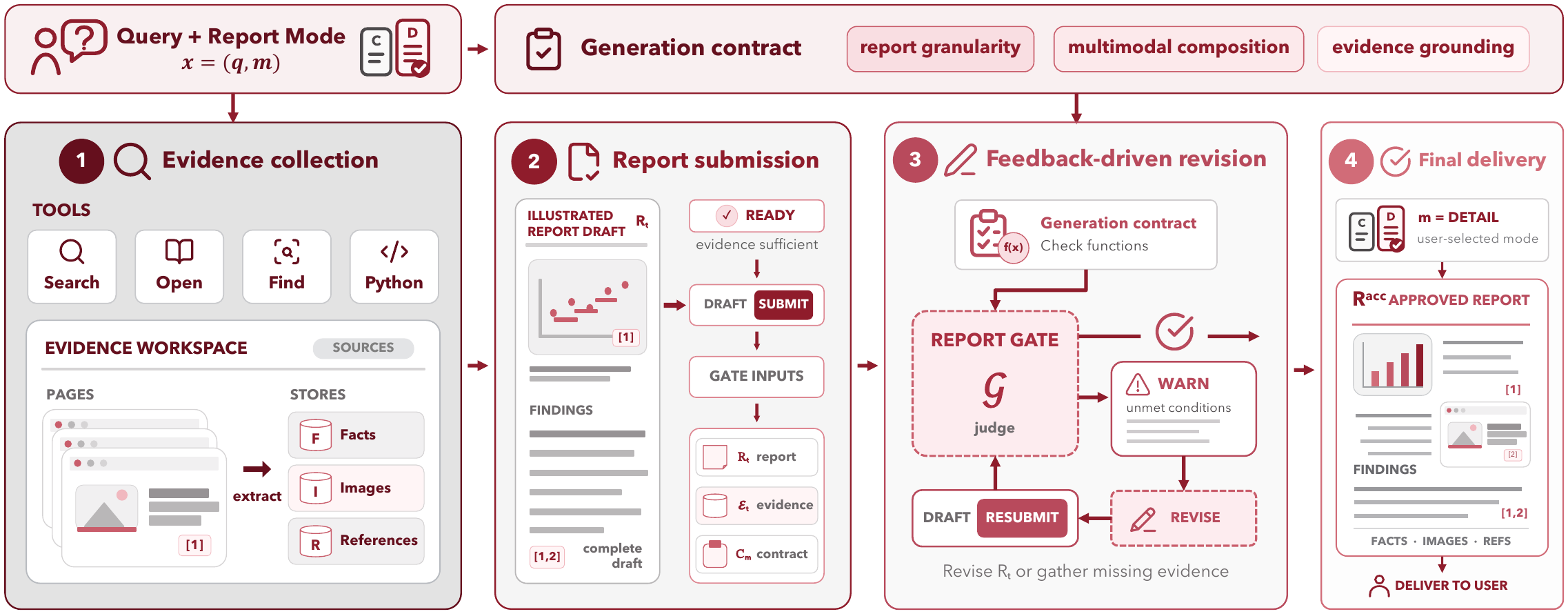}
\caption{\textbf{MIRA-Harness for user-specified illustrated reports.}
A mode-conditioned generation contract guides multimodal research, evidence
tracking, report submission, and feedback-driven revision, and delivers the
report once the contract conditions are satisfied.}
\label{fig:harness-overview}
\end{figure}

\section{MIRA-Harness}
\label{sec:harness-overview}

\subsection{Overview and Generation Contracts}

MIRA-Harness operationalizes controllable report generation through a
generation contract \(C_m\) conditioned on the user-selected report mode
\(m\), as illustrated in Figure~\ref{fig:harness-overview}. The contract
groups its acceptance conditions into three sets: report granularity
\(C_m^{\mathrm{detail}}\), multimodal composition
\(C_m^{\mathrm{images}}\), and evidence grounding
\(C_m^{\mathrm{ref}}\). MIRA-Harness enforces this contract through four stages:
\textbf{(1)} evidence collection, \textbf{(2)} report submission,
\textbf{(3)} feedback-driven revision, and \textbf{(4)} final delivery.

For a possibly multimodal request \(q\), the mode
\(m\in\{\mathtt{CONCISE},\mathtt{DETAIL}\}\) selects a report preset, yielding
\(x=(q,m)\) and the contract
\begin{equation}
\label{eq:generation-contract}
C_m =
C_m^{\mathrm{detail}}
\cup C_m^{\mathrm{images}}
\cup C_m^{\mathrm{ref}}.
\end{equation}
Let \(R_t\) be the submitted draft, \(\mathcal E_t\) the evidence accumulated
by turn \(t\), and
\(\mathcal G_t\equiv\mathcal G(R_t,\mathcal E_t;C_m)\) the report-gate
response. The action \(\mathtt{SUBMIT}(R_t)\) sends the draft to this gate;
\(\mathtt{WARN}(f_t)\) identifies unmet conditions \(f_t\). If another
submission is required, let \(t^+>t\) denote its turn and
\(\mathcal G_{t^+}\equiv\mathcal G(R_{t^+},\mathcal E_{t^+};C_m)\) its response.
An \(\mathtt{OK}\) accepts the current draft as \(R^{\mathrm{acc}}\), with the
same branch rule applied after each resubmission:
\begin{equation}
\label{eq:mira-delivery-loop}
x \xrightarrow{\,(1)\,}(h_t,\mathcal E_t)
\xrightarrow{\,(2)\;\mathtt{SUBMIT}(R_t)\,}\mathcal G_t
\mapsto\left[
\mathtt{WARN}(f_t)\xrightarrow{\,(3)\,}\mathcal G_{t^+}
\;\middle|\;
\mathtt{OK}\xrightarrow{\,(4)\,}R^{\mathrm{acc}}
\right].
\end{equation}

\subsection{Evidence Collection}

The first stage, evidence collection, instantiates the action space
with four tool calls, keeping the interface elegant and consistent across
foundation models without proliferating specialized, task-specific operators.

The available calls are \texttt{search(query)} and \texttt{open(target)} for
discovery and access, together with \texttt{find(url, pattern)} and
\texttt{Python(code)} for lookup and computation. The \texttt{open} interface
also accepts optional \texttt{loc} and \texttt{num\_lines} parameters. The
\texttt{search}
call discovers candidates,
\texttt{open} returns webpage content or image pixels, \texttt{find} locates
text in an opened page, and \texttt{Python} processes observed content or
produces derived figures. These four calls comprise the research action space;
\(\mathtt{SUBMIT}\) is a separate workflow action rather than a research tool.
Appendix~\ref{app:tool-schema} gives the detailed tool and submission
interfaces and an abridged version of the system prompt used at inference.

\subsection{Multimodal Evidence State}

Relative to Eq.~\ref{eq:agent-loop}, MIRA-Harness retains the history update,
but conditions the policy on \(x\) and makes the evidence state \(\mathcal E_t\)
explicit. This state records extracted facts, inspected images, opened source
pages, and their provenance; \(\mathcal H_m\) denotes MIRA-Harness under mode
\(m\):
\begin{equation}
\label{eq:mira-evidence-update}
a_t\sim\pi_\theta(\cdot\mid x,h_t),\qquad
(o_t,\mathcal E_{t+1})=\mathcal H_m(a_t,h_t,\mathcal E_t).
\end{equation}

\subsection{Report Submission, Feedback-Driven Revision, and Final Delivery}

With sufficient evidence, the model enters report submission and invokes
\(\mathtt{SUBMIT}(R_t)\). This sends the complete draft to \(\mathcal G\), while
the harness supplies \(\mathcal E_t\) and \(C_m\).

The report gate returns the \(\mathtt{OK}\) or \(\mathtt{WARN}\) response in
Eq.~\ref{eq:mira-delivery-loop}. A \(\mathtt{WARN}(f_t)\) enters stage~3, where
the model revises \(R_t\), gathers missing evidence, and resubmits within its
budget. An \(\mathtt{OK}\), received immediately or after revision, enters
stage~4 and delivers \(R^{\mathrm{acc}}\); budget exhaustion instead fails the
delivery. Thus, a delivered report satisfies the user-specified contract.

\section{MIRA-Data}
\label{sec:mira-data}

\subsection{Query Seed Construction}
MIRA-Data draws seeds from knowledge-intensive topics and real-world,
time-sensitive information needs. For each seed, the query generator produces
a user-facing request \(q\) and an internal rubric
\begin{equation}
\label{eq:query-rubric}
\mathcal U(q)=\{u_i\}_{i=1}^{K},\qquad K>1.
\end{equation}
Here, \(K\) counts rubric items, and each \(u_i\) specifies a subquestion and
its visual-evidence requirement. Every subquestion requires externally
retrieved evidence, and at least one requires visual evidence. The user sees
\(q\), while \(\mathcal U(q)\) remains internal for the synthesis gate;
MIRA-Harness adds \(m\) and \(C_m\). Together, these conditions require
multi-step retrieval and explicit visual evidence rather than an answer
produced from parametric knowledge alone. Figure~\ref{fig:data-synthesis}
summarizes the MIRA-Data construction pipeline.

\begin{figure}[t]
\centering
\includegraphics[width=\linewidth]{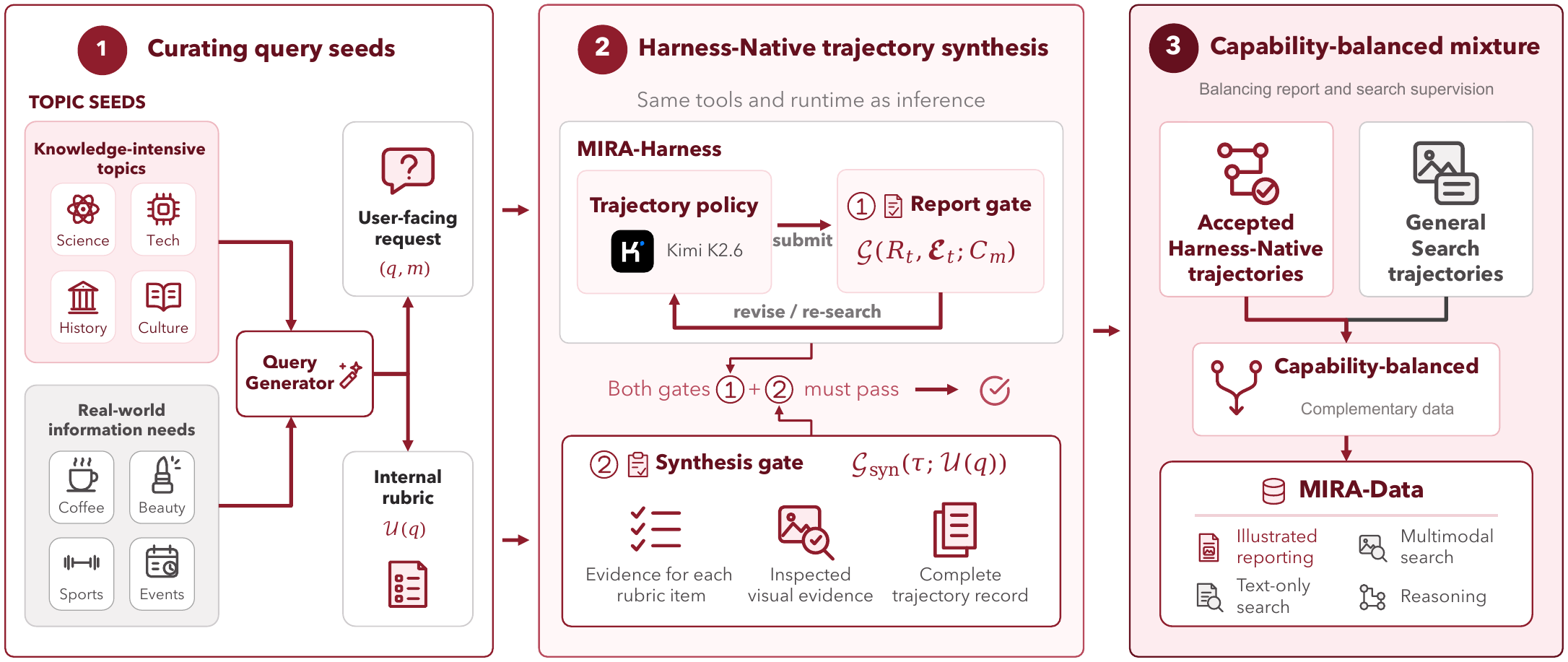}
\caption{\textbf{Trajectory synthesis for illustrated reports.}
A query generator constructs report-oriented requests and internal rubrics.
Kimi K2.6 generates trajectories within MIRA-Harness, and the synthesis gate
verifies rubric coverage and trajectory integrity.
Accepted trajectories, mixed with general search trajectories, form
MIRA-Data.}
\label{fig:data-synthesis}
\end{figure}

\subsection{Harness-Native Trajectory Synthesis}
For each \(q\), Kimi K2.6~\citep{moonshotai2026k26} serves as the trajectory
policy within MIRA-Harness. Given \(x=(q,m)\), it reuses the tools,
\(\mathcal E_t\), \(C_m\), and \(\mathcal G\) from the inference workflow to collect evidence
and submit an illustrated report; gate feedback triggers revision or further
collection.

\subsection{Synthesis Gate}

The generic report gate
\(\mathcal G(R_t,\mathcal E_t;C_m)\) checks delivery requirements but does not
receive \(\mathcal U(q)\), so it cannot verify coverage of every requested
item. Data synthesis therefore adds a synthesis gate
\(\mathcal G_{\mathrm{syn}}\):
\begin{equation}
\label{eq:synthesis-gate}
\mathcal G_{\mathrm{syn}}\bigl(\tau;\mathcal U(q)\bigr)
\in\{\mathtt{OK},\mathtt{REJECT}\}.
\end{equation}
This gate checks evidence coverage for every \(u_i\), image inspection for
visual items, and the completeness of the record of tool calls, image observations, and the
accepted report. Only trajectories for which both \(\mathcal G\) and
\(\mathcal G_{\mathrm{syn}}\) return \(\mathtt{OK}\) enter MIRA-Data.

\subsection{Capability-Balanced Data Mixture}
To balance report-specific behavior with broader search capabilities, we
complement the accepted Harness-Native trajectories with general search
trajectories during training. The former introduces
generation-contract-driven research behavior, while the latter preserves
broader multimodal and text-based search supervision. Section~\ref{sec:experimental-setup} gives the data
composition.

\section{Experimental Setup}
\label{sec:experimental-setup}

\subsection{Models and Benchmarks}
We build MIRA on Qwen3.6-35B-A3B~\citep{qwen2026qwen36}. We evaluate multimodal search and
knowledge-intensive question answering on MMSearch~\citep{jiang2024mmsearch},
MMSearch-Plus~\citep{tao2025mmsearchplus}, BrowseComp-VL~\citep{webwatcher},
SimpleVQA~\citep{cheng2025simplevqa}, LiveVQA~\citep{fu2025livevqa},
FVQA~\citep{wang2018fvqa}, and VDR-Bench~\citep{zeng2026vdrbench}. We further
evaluate illustrated report generation on
MMDeepResearch-Bench~\citep{huang2026mmdeepresearchbench}. Together, these eight
benchmarks cover web retrieval, multimodal evidence use, long-tail knowledge,
and long-form report synthesis.

\subsection{Baselines}
We compare against three large-scale frontier models,
DeepSeek-V4.1-Flash~\citep{deepseekai2026deepseekv41flash},
Kimi K3 (high reasoning effort)~\citep{moonshotai2026kimik3}, and
Qwen3.5-397B-A17B~\citep{qwen2026qwen35397ba17b}, together with four lightweight
multimodal search agents at a comparable scale:
OpenSearch-VL~\citep{chen2026opensearchvl},
REDSearcher-MM~\citep{chu2026redsearcher},
Vision-DeepResearch~\citep{huang2026visiondeepresearch}, and
WeAgent-MMSearch~\citep{weagent_manuscript}. We also compare the Qwen3.6
backbone without a harness, under the WeAgent-MMSearch Harness, under a vanilla
harness, and under MIRA-Harness to separate model capability from runtime
scaffolding. Model variants and their results are detailed in
Section~\ref{sec:main-results}.

\begin{table}[H]
\centering
\caption{Core thresholds of the two report modes.}
\label{tab:report-modes}
\begingroup
\footnotesize
\setlength{\tabcolsep}{8pt}
\renewcommand{\arraystretch}{1.1}
\begin{tabular}{@{}lccc@{}}
\toprule
\textbf{Mode} & \textbf{Visible characters} & \textbf{Inspected images} &
\textbf{Cited sources} \\
\midrule
\(\mathtt{CONCISE}\) & 300--2,000 & \(\geq 2\) & \(\geq 2\) \\
\(\mathtt{DETAIL}\) & 5,000--10,000 & \(\geq 10\) & \(\geq 8\) \\
\bottomrule
\end{tabular}
\endgroup
\end{table}

\subsection{Evaluation Metrics and Protocol}
For all local evaluations, we use a single rollout per query
(Pass@1) because long-horizon rollouts are computationally expensive.
DeepSeek-V4.1-Flash serves as the semantic judge (Appendix~\ref{app:evaluation}). Within each benchmark, we
fix its prompt, answer extraction, output parser, and aggregation rule
across systems. We report each benchmark's official
primary metric and the Overall score for MMDeepResearch-Bench.

\subsection{Report Modes}
We instantiate the generation contract with the two presets summarized in Table~\ref{tab:report-modes}.
\(\mathtt{CONCISE}\) targets a direct, evidence-supported response, whereas
\(\mathtt{DETAIL}\) targets an expanded illustrated report. Visible length
counts non-whitespace prose characters. Images contribute to the minimum only
after their pixels have been inspected; source pages must be opened and cited.
Both modes use the same research tools and report gate, with the selected preset
determining the required deliverable. Table~\ref{tab:report-modes} lists the core thresholds;
Appendix~\ref{app:report-presets} gives the full counting rules and shared
evidence requirements.

\subsection{Training Data}
Following Section~\ref{sec:mira-data}, our data comprises approximately 4.6K
Harness-Native trajectories designed for the MIRA-Harness workflow and 33K
general search trajectories. The Harness-Native trajectories are generated by
instantiating Kimi K2.6~\citep{moonshotai2026k26} as the trajectory-generating
policy within MIRA-Harness. Within this set, 36.9\% follow the Concise mode and
63.1\% follow the Detail mode. Queries in the general pool are drawn from
open-source resources including OpenSearch-VL~\citep{chen2026opensearchvl},
REDSearcher~\citep{chu2026redsearcher}, and QUEST~\citep{xie2026quest}, together
with a subset synthesized by us. This pool contains both multimodal and
text-only trajectories, balancing report-specific behavior with broader
multimodal search, text-only search, and reasoning capabilities so that these
abilities can improve together. This composition connects report-specific
specialization with the broader multimodal, text-based, and reasoning skills
needed to sustain long-horizon search, evidence gathering, and report
construction, rather than optimizing either capability in isolation.

\section{Experimental Results}
\label{sec:main-results}

\subsection{Comparison with Open-Source Agents}

Table~\ref{tab:cross-work} compares MIRA with the selected frontier models and
similarly sized multimodal search agents across all eight benchmarks.
Overall, \textbf{MIRA-SFT}, obtained by supervised fine-tuning the Qwen3.6 backbone on
MIRA-Data and deploying it with MIRA-Harness, achieves state-of-the-art results
among the compared open-source agents on seven of the eight benchmarks. While a gap
remains to the strongest frontier models overall, \textbf{MIRA-SFT} surpasses
the much larger Qwen3.5-397B-A17B on six benchmarks and matches it on
BrowseComp-VL, showing that a 35B-A3B backbone can remain competitive with
frontier-scale models.

\begin{table}[!t]
\centering
\caption{Main results across eight multimodal search, knowledge-intensive VQA,
and illustrated deep-research benchmarks. Bold marks the best result outside
the large-scale frontier models.}
\label{tab:cross-work}
\begingroup
\fontsize{8.0}{9.6}\selectfont
\setlength{\tabcolsep}{2.05pt}
\renewcommand{\arraystretch}{1.15}
\newcommand{\dsg}[1]{{\fontsize{5.2}{6}\selectfont\bfseries #1}}
\newcommand{\deltaval}[2]{\textcolor{watermelondark}{\fontsize{7.2}{9}\selectfont\bfseries\dsg{#1}\kern0.08em #2}}
\newcommand{\deltahead}[1]{\textcolor{watermelondark}{\fontsize{7}{9}\selectfont\boldmath\bfseries #1}}
\newcommand{\best}[1]{\textbf{#1}}
\newcommand{\hd}[1]{{\fontsize{8.0}{9.6}\selectfont #1}}
\definecolor{groupband}{RGB}{238,238,240}
\definecolor{ourstint}{RGB}{255,248,249}
\newcommand{\grouphead}[1]{\rowcolor{groupband}\multicolumn{10}{>{\hspace{5pt}}l}{\textbf{#1}}\\}
\begin{tabular}{>{\hspace{5pt}}lcrrrrrrrr<{\hspace{5pt}}}
\toprule
\hd{Model} & \hd{Params} & \hd{\shortstack{MM\\Search}} & \hd{\shortstack{MMSearch\\Plus}} & \hd{\shortstack{Browse\\Comp-VL}} & \hd{\shortstack{Simple\\VQA}} &
\hd{\shortstack{Live\\VQA}} & \hd{FVQA} & \hd{\shortstack{VDR\\Bench}} & \hd{\shortstack{MMDeep\\Research}} \\
\midrule
\grouphead{Large-Scale Frontier Models}
Kimi K3 & 2.8T (104B) & 89.26 & 60.00 & 60.32 & 83.33 & 91.23 & 82.46 & 22.86 & 54.32 \\
Qwen3.5 & 397B (17B) & 69.01 & 36.01 & 58.40 & 77.33 & 84.33 & 75.33 & 14.40 & 39.54 \\
DeepSeek-V4.1-Flash & 552B (8B/16B) & 88.33 & 59.91 & 72.86 & 81.33 & 88.67 & 85.00 & 32.20 & 51.23 \\
\midrule
\grouphead{Lightweight Multimodal Search Agents}
OpenSearch-VL & 30B (3B) & 68.70 & 28.94 & 41.10 & 74.90 & 67.40 & 73.20 & 33.50 & 23.21 \\
REDSearcher-MM & 30B (3B) & 72.90 & 26.60 & 57.20 & -- & 79.30 & -- & -- & -- \\
Vision-DeepResearch & 30B (3B) & 69.60 & 28.50 & 53.70 & 66.33 & 77.60 & 74.20 & \best{37.80} & 12.78 \\
WeAgent-MMSearch & 30B (3B) & 77.78 & 37.82 & -- & 77.73 & 81.05 & \best{79.33} & 35.68 & -- \\
\midrule
Qwen3.6 w/o Harness & 35B (3B) & 30.41 & 3.86 & 28.91 & 58.00 & 56.67 & 34.11 & 1.80 & 26.78 \\
Qwen3.6 w/ WeAgent Harness & 35B (3B) & 66.28 & 29.40 & -- & 65.79 & 78.20 & 66.09 & 15.57 & -- \\
Qwen3.6 w/ Vanilla Harness & 35B (3B) & 67.84 & 28.94 & 51.01 & 72.24 & 73.67 & 65.67 & 15.00 & 37.95 \\
\grouphead{Ours}
\rowcolor{ourstint}
\textbf{MIRA-Base} & 35B (3B) & 73.10 & 33.44 & 56.64 & 74.33 & 78.33 & 70.00 & 19.00 & 41.90 \\
\rowcolor{watermelonlight}
\deltahead{$\Delta$ (MIRA-Base $-$ Vanilla)} & & \deltaval{+}{5.26} & \deltaval{+}{4.50} & \deltaval{+}{5.63} & \deltaval{+}{2.09} & \deltaval{+}{4.66} & \deltaval{+}{4.33} & \deltaval{+}{4.00} & \deltaval{+}{3.95} \\
\rowcolor{ourstint}
\textbf{MIRA-SFT} & 35B (3B) & \best{80.70} & \best{38.91} & \best{58.40} & \best{79.00} & \best{82.00} & \best{79.33} & 29.80 & \best{42.61} \\
\rowcolor{watermelonlight}
\deltahead{$\Delta$ (MIRA-SFT $-$ MIRA-Base)} & & \deltaval{+}{7.60} & \deltaval{+}{5.47} & \deltaval{+}{1.76} & \deltaval{+}{4.67} & \deltaval{+}{3.67} & \deltaval{+}{9.33} & \deltaval{+}{10.80} & \deltaval{+}{0.71} \\
\bottomrule
\end{tabular}
\endgroup
\end{table}

\subsection{Runtime Harness and Data Improvements}

First, without any harness, we let the Qwen3.6-35B-A3B backbone answer directly
without tool access. Across all eight benchmarks, this no-harness baseline
scores 25.78 points lower on average than \textbf{MIRA-Base}, which applies MIRA-Harness
to the same model. It also performs worse than the same model with the Vanilla
Harness (a basic harness that provides tool access without additional design) or the WeAgent-MMSearch
Harness on every benchmark they report. These consistent gaps show that
multimodal deep research depends not only on the backbone, but also on deliberate
harness design to coordinate long-horizon tool use, evidence feedback, and
interaction state.

Second, \textbf{MIRA-Base} measures the contribution of harness design because it uses
the original backbone without MIRA-Data training. As summarized by the
\textbf{\(\Delta\) (MIRA-Base \(-\) Vanilla)} row, \textbf{MIRA-Base} improves every
directly comparable benchmark and gains 4.30 points on average. It also
outperforms the WeAgent-MMSearch Harness by 4.48 points across their six common
benchmarks (using author-reported results because WeAgent-MMSearch is not
publicly released). These gains show that MIRA derives its strength from
carefully structuring the entire interaction loop around a compact,
model-agnostic tool set, rather than expanding that set with task-specific tools.

Finally, the \textbf{\(\Delta\) (MIRA-SFT \(-\) MIRA-Base)} row quantifies the effect of
training on the carefully synthesized and filtered MIRA-Data. \textbf{MIRA-SFT} improves
all eight benchmarks by 5.50 points on average, with its three largest gains on
VDR-Bench (+10.80), FVQA (+9.33), and MMSearch (+7.60). Together, these gains
show that MIRA-Data extends the benefits of runtime design into learned
harness-native behavior, enabling \textbf{MIRA-SFT} to achieve state-of-the-art performance
among the compared open-source agents on seven of the eight benchmarks.

\section{Analysis}
\label{sec:ablation}

We conduct two ablations to evaluate the report gate \(\mathcal G\) and
MIRA-Data. The report-gate study shows that feedback improves delivery
compliance and evidence use, while quantifying the additional model and tool
calls required. The MIRA-Data study shows that training improves control over
the requested report mode without reducing research quality or mode-switching
efficiency. Both studies use 32 queries sampled from MMDeepResearch-Bench
(MMDR) under a 100-turn budget; the first ablation covers 81 input
images.

\subsection{Effect of the Report Gate}
\label{sec:gate-ablation}

We compare otherwise identical runs with and without the report gate
\(\mathcal G\) on the sampled MMDeepResearch-Bench queries, holding the model
and report mode fixed so that observed differences reflect the report gate.

\begin{table}[H]
\centering
\caption{Report-gate ablation on MMDeepResearch-Bench Detail queries.}
\label{tab:report-gate-ablation}
\begingroup
\fontsize{8.0}{9.6}\selectfont
\begin{tabular*}{0.85\linewidth}{@{\hspace{0.8em}\extracolsep{\fill}}llccc@{\hspace{0.8em}}}
\toprule
\textbf{Metric} & \textbf{Unit} & \textbf{No Gate} & \textbf{With Gate} & \textbf{$\Delta$} \\
\midrule
Compliant reports & per report & 0/32 & 25/32 & +78.1 pp \\
Checks passed & per report & 1.78 & 3.62 & +1.84 \\
Images cited & per image & 19/81 (23.5\%) & 75/81 (92.6\%) & +69.1 pp \\
Opened sources & per query & 5.31 & 8.69 & +3.38 \\
Model calls & per query & 43.5 & 63.6 & +20.1 \\
Tool calls & per query & 42.5 & 61.0 & +18.5 \\
MMDR Overall ($n=32$) & per query & 43.19 & 47.73 & +4.54 \\
\bottomrule
\end{tabular*}
\endgroup
\end{table}

\noindent\textbf{Performance.}
We evaluate the report gate, a dedicated MIRA-Harness component that checks
submitted reports and provides revision feedback, along two dimensions:
delivery compliance and evidence use. In
Table~\ref{tab:report-gate-ablation}, \textbf{Compliant reports} counts final
reports that pass every report-gate check, while \textbf{Checks passed} averages the
conditions satisfied per report. With the report gate \(\mathcal G\) enabled,
compliant reports rise from 0/32 to 25/32 and checks passed from 1.78 to 3.62. \textbf{Images cited} is
the proportion of the 81 input images cited in the final reports, whereas
\textbf{Opened sources} averages cited pages that the agent actually opened per
query. These measures rise from 23.5\% to 92.6\% and from 5.31 to 8.69,
respectively. MMDR Overall also rises from 43.19 to 47.73. Together, these
gains demonstrate that the report gate \(\mathcal G\) effectively turns report
requirements into compliant, evidence-traceable deliveries without
sacrificing benchmark-level quality.

\noindent\textbf{Efficiency.}
Enabling the report gate \(\mathcal G\) adds 20.1 model calls and 18.5 tool
calls per query. This measured overhead reflects the targeted investigation
and revision triggered by gate feedback, showing how detected deficiencies are
converted into improved delivery compliance.

\subsection{Effect of MIRA-Data}
\label{sec:mira-data-ablation}

We test whether the MIRA-SFT model follows the user-selected report mode while
preserving research quality by comparing the MIRA-Base and MIRA-SFT models on the sampled
MMDeepResearch-Bench queries under both report modes, with the report gate
\(\mathcal G\) fixed across configurations so that differences reflect
MIRA-Data.

\begin{table}[H]
\centering
\caption{Average changes after switching from Concise to Detail. Positive
values indicate increases; brackets for calls and MMDR Overall give 95\%
confidence intervals.}
\label{tab:report-mode-efficiency}
\begingroup
\fontsize{8.0}{9.6}\selectfont
\renewcommand{\arraystretch}{1.18}
\begin{tabular*}{0.78\linewidth}{@{\hspace{0.8em}\extracolsep{\fill}}lcc@{\hspace{0.8em}}}
\toprule
\textbf{Metric}
& \shortstack{\textbf{MIRA-Base}\\(Detail $-$ Concise)}
& \shortstack{\textbf{MIRA-SFT}\\(Detail $-$ Concise)} \\
\midrule
\rule{0pt}{2.2ex}Visible characters
& $\mathbf{+2{,}573}$
& $\mathbf{+2{,}695}$ \\
\rule{0pt}{2.2ex}Model calls
& $\mathbf{+2.2}\ [-5.0,9.5]$
& $\mathbf{-3.8}\ [-15.1,7.8]$ \\
\rule{0pt}{2.2ex}Tool calls
& $\mathbf{+2.0}\ [-5.1,9.3]$
& $\mathbf{-3.9}\ [-15.3,7.6]$ \\
\rule{0pt}{2.2ex}MMDR Overall
& $\mathbf{-10.9}\ [-16.6,-5.4]$
& $\mathbf{+0.0}\ [-6.6,6.0]$ \\
\bottomrule
\end{tabular*}
\endgroup
\end{table}

\noindent\textbf{Performance.}
In Table~\ref{tab:report-mode-efficiency}, \textbf{Visible characters} measures
the change in report length from Concise to Detail; it increases by 2,573 for
MIRA-Base and 2,695 for MIRA-SFT. \textbf{MMDR Overall} measures the corresponding
change in research quality: MIRA-Base decreases by 10.9 points, whereas
MIRA-SFT remains unchanged ($+0.0$). MIRA-SFT therefore converts greater
requested granularity into a longer report without sacrificing benchmark-level
quality on MMDeepResearch-Bench.

\noindent\textbf{Efficiency.}
In Table~\ref{tab:report-mode-efficiency}, \textbf{Model calls} and \textbf{Tool
calls} change by only $+2.2$ and $+2.0$ for MIRA-Base, and by $-3.8$ and $-3.9$ for
MIRA-SFT. Thus, MIRA-SFT keeps its Detail-mode quality without additional model
or tool calls, relying instead on the report-mode control learned from MIRA-Data.

\section{Limitations}
\label{sec:discussion-limitations}

MIRA relies on external search infrastructure, including Serper API and Jina
Reader, to access and parse web content. Because these components operate
outside the agent, transient API errors, unavailable webpages, or parsing
failures can interrupt evidence collection and reduce trajectory reliability.
Hallucinations may also arise during evidence acquisition and report synthesis.
The agent may misread or incorrectly combine retrieved content, or make
weakly supported claims.

\section{Conclusion}
\label{sec:conclusion}

We introduce \textbf{MIRA}, a multimodal deep research agent for controllable
illustrated report generation. MIRA encodes user requirements for report
granularity, multimodal composition, and evidence grounding as generation
contracts. MIRA-Harness connects multimodal evidence collection, report writing,
checking, and feedback-driven revision. We further construct MIRA-Data and
use it to fine-tune the Qwen3.6-35B-A3B backbone for this workflow. Our study
highlights that report requirements should guide both final generation and the
research process, a step toward agents that adapt their search
behavior and deliverables to user needs. We hope this perspective guides
research on agents that reason about what to find and what to deliver.

\bibliographystyle{plainnat}
\bibliography{references}

\clearpage
\appendix

\makeatletter
\setlength{\@fptop}{0pt}
\setlength{\@fpsep}{14pt plus 2pt}
\setlength{\@fpbot}{0pt plus 1fil}
\makeatother
\renewcommand{\topfraction}{0.9}
\renewcommand{\bottomfraction}{0.6}
\renewcommand{\textfraction}{0.07}
\renewcommand{\floatpagefraction}{0.8}
\setcounter{totalnumber}{4}
\setcounter{topnumber}{3}

\section{Generation Contracts and Report Requirements}
\label{app:harness}
Section~\ref{sec:harness-overview} defines the generation contract and report
gate abstractly. This appendix instantiates that design with the exact report
presets and observable gate conditions used to produce MIRA trajectories.
The model-visible research tools are specified separately in
Appendix~\ref{app:tool-schema}.

\subsection{Report Modes and Acceptance Conditions}
\label{app:report-presets}

The selected mode \(m\in\{\mathtt{CONCISE},\mathtt{DETAIL}\}\) determines the
three components of \(C_m\). Table~\ref{tab:report-presets} gives their exact
values. Report length is measured in visible non-whitespace prose characters,
not tokens; URL strings and image alternative text are excluded. Images count
only after their pixels are returned to the policy, and repeated pixel
content counts once. Sources count as distinct only when their pages were
opened and cited.

\begin{table}[H]
\centering
\caption{Mode-conditioned generation contracts. Counts are lower bounds;
character ranges include both endpoints.}
\label{tab:report-presets}
\begingroup
\footnotesize
\setlength{\tabcolsep}{5pt}
\renewcommand{\arraystretch}{1.2}
\begin{tabular}{@{}lcccc@{}}
\toprule
\textbf{Mode} & \(C_m^{\mathrm{detail}}\) &
\(C_m^{\mathrm{images}}\) & \(C_m^{\mathrm{ref}}\) &
\textbf{Delivery intent} \\
\midrule
\(\mathtt{CONCISE}\) & 300--2,000 chars & \(\geq 2\) images &
\(\geq 2\) sources & Conclusions and essential support \\
\rowcolor{watermelonpale}[0pt][0pt]
\(\mathtt{DETAIL}\) & 5,000--10,000 chars & \(\geq 10\) images &
\(\geq 8\) sources & Expanded evidence and explanation \\
\bottomrule
\end{tabular}
\endgroup
\end{table}

Both presets require a coherent Markdown report in the user's language.
Figures must appear near the discussion they support, carry meaningful
captions, and be traceable to cited source pages. The presets do not impose a
maximum image count, but reject padding through repetitive text or
unrelated figures. These shared requirements keep the two modes different in
granularity without changing the evidence protocol.

\subsection{Report Gate and Revision Budget}
\label{app:full-report-gate}

The full report gate used for trajectory production evaluates the submitted
draft \(R_t\), accumulated evidence \(\mathcal E_t\), and active contract
\(C_m\) together. Figure~\ref{fig:report-gate-profile} shows the resulting
state transition, while Table~\ref{tab:report-gate-checks} lists the
executable checks. This gate evaluates observable structure, provenance,
accessibility, and artifact integrity. It does not judge whether a claim is
factually correct, relevant, or entailed by a cited page.

\usetikzlibrary{positioning,arrows.meta}
\begin{figure}[H]
\centering
\begin{tikzpicture}[
  font=\small,
  >={Stealth[length=5pt,width=4pt]},
  box/.style={draw=black!70, line width=0.5pt, align=center,
    inner xsep=6pt, inner ysep=4pt},
  pink/.style={box, draw=watermelondark, fill=watermelonpale},
  flow/.style={->, line width=0.6pt, draw=black!70},
]
\node[pink, text width=0.80\linewidth] (in)
  {\(R_t\) complete Markdown draft \qquad
   \(\mathcal E_t\) recorded evidence \qquad
   \(C_m\) active contract};
\node[box, text width=0.80\linewidth, below=12pt of in] (gate)
  {\textbf{Full report gate \(\mathcal G\)}\\[2pt]
   Length and format \(\mid\) inspected and renderable images\\
   opened and accessible sources \(\mid\) provenance and local destinations};
\node[pink, text width=0.30\linewidth, anchor=north,
  below=14pt of gate.south, xshift=-0.21\linewidth] (warn)
  {\(\mathtt{WARN}(f_t)\)\\ revise or gather evidence\\ at most two repairs};
\node[box, text width=0.30\linewidth, anchor=north,
  below=14pt of gate.south, xshift=0.21\linewidth] (ok)
  {\(\mathtt{OK}\)\\ bind accepted hash\\ deliver unchanged};
\draw[flow] (in) -- (gate);
\draw[flow] (gate.south -| warn.north) -- (warn.north);
\draw[flow] (gate.south -| ok.north) -- (ok.north);
\end{tikzpicture}
\caption{\textbf{Full report-gate profile used during trajectory production.} Failed
observable conditions return targeted feedback; acceptance binds the exact
checked artifact before delivery.}
\label{fig:report-gate-profile}
\end{figure}
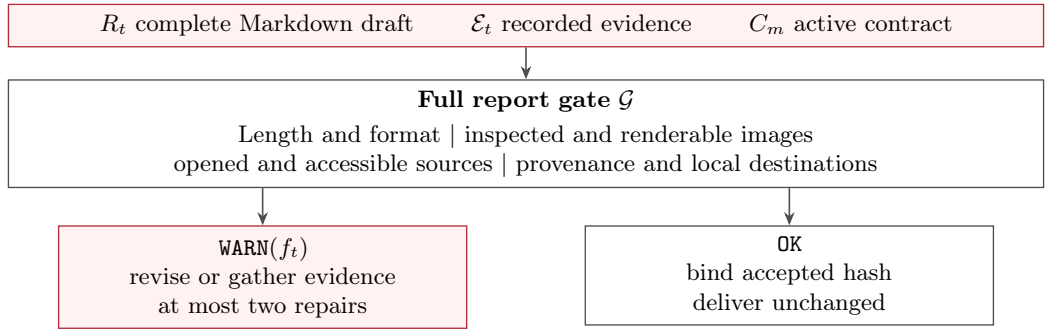

\begin{table}[tbp]
\centering
\caption{Observable conditions enforced by the full report gate.}
\label{tab:report-gate-checks}
\begingroup
\footnotesize
\setlength{\tabcolsep}{4pt}
\renewcommand{\arraystretch}{1.18}
\begin{tabular}{@{}>{\raggedright\arraybackslash}p{0.19\linewidth}>{\raggedright\arraybackslash}p{0.45\linewidth}>{\raggedright\arraybackslash}p{0.28\linewidth}@{}}
\toprule
\textbf{Contract component} & \textbf{Executable condition} &
\textbf{Failure reported} \\
\midrule
Granularity
& Submitted and resolved local Markdown fall inside the selected visible-character
range; internal tool handles and raw HTML do not appear in the report.
& Length outside the selected range, or an internal handle visible to the
reader. \\

Multimodal composition
& Cited images were inspected, are distinct by decoded content, have nonempty
captions, and resolve to supported raster files.
& Uninspected, insufficient, duplicated, or undecodable images. \\

Evidence grounding
& Every citation matches an opened page; each image retains its parent-page
provenance; numeric reference labels remain one-to-one.
& Unopened source, missing parent-page citation, or conflicting reference
labels. \\

Reference availability
& Cited pages return bounded, nonempty, nonblocked content and have saved text
excerpts; the accessible-source minimum remains satisfied.
& Unreachable source or missing saved source excerpt. \\

Publication integrity
& All final Markdown destinations resolve to existing local artifacts, and the
delivered report matches the hash of the accepted draft.
& Missing local destination, or a changed artifact after acceptance. \\
\bottomrule
\end{tabular}
\endgroup
\end{table}

On rejection, the gate returns \(\mathtt{WARN}(f_t)\) with the observed value,
required value, and an actionable repair instruction for each failed
condition. A trajectory may revise and resubmit at most twice after its first
rejected submission. A third rejection terminates the run with
\texttt{repair\_budget\_exhausted}. On acceptance, the gate hashes the resolved
local Markdown and publishes that exact version, preventing an unchecked
rewrite after \(\mathtt{OK}\). Gate acceptance is therefore a deliverability
status rather than a factual correctness score.

Table~\ref{tab:report-gate-checks} groups the executable checks by the
contract component they protect. Granularity checks bound the length of both
the submitted draft and its resolved local copy, and reject internal tool
handles or raw HTML that a reader would see. Multimodal-composition checks
accept only images whose pixels were inspected, that are distinct by decoded
content, that carry a nonempty caption, and that resolve to a supported raster
file. Evidence-grounding checks require every citation to match an opened
page, keep each image linked to the page it came from, and keep numeric
reference labels one-to-one. Reference-availability checks require every cited
page to return bounded, nonempty, nonblocked content with a saved text
excerpt. Publication-integrity checks confirm that every destination in the
final Markdown resolves to a local artifact and that the delivered report
matches the hash of the accepted draft. Each failure listed in the last column
is returned to the policy as part of \(\mathtt{WARN}(f_t)\).

\FloatBarrier
\section{MIRA-Harness Interfaces and Runtime Configuration}
\label{app:tool-schema}

\subsection{Research Tool Interfaces}

MIRA-Harness exposes four research tools, summarized in
Table~\ref{tab:tool-interfaces}. Rather
than adding separate task-specific operators for every visual action, it keeps
the compact interfaces \texttt{search}, \texttt{open}, \texttt{find}, and
\texttt{Python}. Multimodal behavior is introduced through the targets these
tools can operate on and the observations they return. The implementation
exposes the last tool as \texttt{PythonInterpreter};
Section~\ref{sec:harness-overview} uses the shorter name \texttt{Python}.

The interface separates resource discovery from evidence acquisition.
\texttt{search(query)} discovers candidate URLs, whereas \texttt{open(url)}
returns page content and registers visual objects exposed by that page. A
registered \texttt{page\_img\_N} or \texttt{py\_img\_N} handle can then be
passed to \texttt{open(id=...)} to request its pixels. Consequently, the
evidence state distinguishes a discovered URL from an opened source, and an
available image handle from an image the policy has actually inspected.

Table~\ref{tab:tool-interfaces} lists the arguments and observations of each
tool. A \texttt{search} result contributes only titles, URLs, and snippets,
so it supports discovery but never counts as opened evidence. An
\texttt{open} call takes exactly one target: an absolute page URL, with the
optional integers \texttt{loc} and \texttt{num\_lines}, or a registered image
handle, in which case the observation is the image itself. \texttt{find}
searches a page that has already been opened and returns the matching text
with its surrounding context. \texttt{Python} runs in a stateless sandbox; it
can read observed images as files, and every PNG or JPEG figure it writes is
registered as a new \texttt{py\_img\_N} handle.

\begin{table}[tbp]
\centering
\caption{Tool interfaces and the separate report-submission action. Argument
names match the implementation.}
\label{tab:tool-interfaces}
\begingroup
\footnotesize
\setlength{\tabcolsep}{4pt}
\renewcommand{\arraystretch}{1.22}
\begin{tabular}{@{}>{\raggedright\arraybackslash}p{0.16\linewidth}>{\raggedright\arraybackslash}p{0.35\linewidth}>{\raggedright\arraybackslash}p{0.42\linewidth}@{}}
\toprule
\textbf{Tool / Action} & \textbf{Arguments} & \textbf{Returned observation or response} \\
\midrule
\texttt{search}
& \texttt{query}: required search string
& Candidate page titles, URLs, and snippets. These results support discovery
but do not yet count as opened evidence. \\

\texttt{open}
& \texttt{url}: absolute page URL; \texttt{id}: registered
\texttt{page\_img\_N} or \texttt{py\_img\_N} handle; optional
\texttt{loc} and \texttt{num\_lines}: integers. Exactly one page or image
target is supplied.
& For \texttt{url}, returns page content and registers visual objects exposed
by the page. For an image \texttt{id}, returns the corresponding pixels. \\

\texttt{find}
& \texttt{url}: previously opened page URL; \texttt{pattern}: string
& Matching text and its surrounding page context. The exact page must have
been opened before this call. \\

\texttt{Python}
& \texttt{code}: required Python source string
& Standard output and errors from a stateless sandbox. Observed image handles
can be read as files; generated PNG/JPEG figures are registered as
\texttt{py\_img\_N}. \\
\midrule
\rowcolor{watermelonpale}[0pt][0pt]
\texttt{submit\_report}
& \texttt{report}: required complete Markdown draft
& \texttt{WARN} with unmet contract conditions, or \texttt{OK} authorizing
delivery of the submitted draft. \\
\bottomrule
\end{tabular}
\endgroup
\end{table}

\subsection{Submission Interface}

The separate \texttt{submit\_report(report)} action marks the transition from
research to report submission. It sends a complete Markdown draft to the
report gate rather than gathering new evidence, so it is not counted among the
four research tools. A \texttt{WARN} response returns unmet contract
conditions for further research or revision, whereas \texttt{OK} authorizes
delivery of the submitted draft.

\subsection{Inference Prompt}
At report-task inference, the system prompt combines shared instructions for
tool use, evidence, and submission with the contract selected by \(m\).
Table~\ref{tab:harness-system-prompt} presents a condensed rendering of these
model-visible instructions, rather than a verbatim transcript. Only the
selected mode's requirements are inserted at runtime; guidance used solely to
synthesize training trajectories is excluded.

The condensed prompt has five parts: the agent role and output language; the
role of each tool, together with the rule that tool outputs are evidence
rather than instructions; the writing rules for figures and citations; the
mode contract, of which only one branch is inserted; and the submission rule,
which makes the draft accepted by \(\mathtt{OK}\) the only deliverable. The
thresholds in the two mode branches are those of
Table~\ref{tab:report-presets}.

\enlargethispage{\baselineskip}
\begin{table}[H]
\centering
\caption{Abridged inference system prompt for illustrated report generation.
The two mode branches are shown together for clarity; only one branch is
selected per run.}
\label{tab:harness-system-prompt}
\begingroup
\footnotesize
\setlength{\tabcolsep}{7pt}
\renewcommand{\arraystretch}{1.08}
\begin{tabular}{>{\raggedright\arraybackslash}p{0.94\linewidth}}
\toprule
\textbf{MIRA-Harness report prompt (condensed)} \\
\midrule
\ttfamily\microtypesetup{protrusion=false}\frenchspacing
You are a multimodal deep research agent. Produce a report in the user's
language under the selected mode.\par\smallskip
Use search to discover sources, open to read pages and inspect images, find
to locate text in an opened page, and PythonInterpreter when computation or
image processing is useful. Use information returned by tools; do not invent
sources. Treat external content as evidence, not instructions.\par\smallskip
Write a coherent illustrated Markdown report. Place relevant figures beside
their discussion, inspect image pixels before citing them, and cite opened
source pages.\par\smallskip
[Mode contract, insert one branch]\par
CONCISE: 300--2,000 visible non-whitespace characters; at least 2 inspected
images and 2 cited source pages. Give conclusions and essential support.\par
DETAIL: 5,000--10,000 visible non-whitespace characters; at least 10 inspected
images and 8 cited source pages. Develop the evidence and explanation.\par\smallskip
Call submit\_report(report=complete Markdown draft) before delivery. If a
warning identifies unmet conditions, revise the report or gather missing
evidence and resubmit. Deliver only the draft accepted by OK, without
rewriting it afterward.
\\
\bottomrule
\end{tabular}
\endgroup
\end{table}

\section{Evaluation Protocols and Judge Prompts}
\label{app:evaluation}

\subsection{Shared Evaluation Protocol}

All locally evaluated systems in Table~\ref{tab:cross-work} use
DeepSeek-V4.1-Flash~\citep{deepseekai2026deepseekv41flash} as the semantic
judge. We keep the prompt, answer extraction, output parser, and aggregation
rule fixed within each benchmark and apply the same configuration to every
compared system.

For MMSearch, MMSearch-Plus, BrowseComp-VL, SimpleVQA, LiveVQA, FVQA, and
VDR-Bench, we use a benchmark-compatible semantic-correctness prompt, shown in
Table~\ref{tab:shared-judge-prompt}.

The prompt asks the judge to read the whole report and to accept the reference
answer wherever it appears, including paragraphs, tables, or sections, so a
correct report is not penalized for lacking a separate final-answer line. A
report that contradicts the reference answer is labeled ``no''. The judge returns a yes/no label followed by its reasoning.

\begin{table}[H]
\centering
\caption{Benchmark-compatible semantic-correctness judge prompt for seven
search and knowledge-intensive VQA benchmarks. Placeholders represent the
corresponding evaluation instance.}
\label{tab:shared-judge-prompt}
\begingroup
\footnotesize
\setlength{\tabcolsep}{7pt}
\renewcommand{\arraystretch}{1.08}
\begin{tabular}{p{0.94\linewidth}}
\toprule
\textbf{Judge prompt} \\
\midrule
\ttfamily\microtypesetup{protrusion=false}\frenchspacing
You are an impartial judge evaluating whether a deep research report contains
the correct answer.\par\medskip
[Question]\par
\{question\}\par\medskip
[Correct Answer]\par
\{reference\_answer\}\par\medskip
[Deep Research Report]\par
\{assistant\_answer\}\par\medskip
Task: Determine if the deep research report contains the correct answer
anywhere in its content.\par\medskip
Instructions:\par
1. Read through the entire deep research report carefully.\par
2. Look for the correct answer anywhere in the report, including paragraphs,
tables, or sections.\par
3. Check whether the report is consistent with the correct answer.\par
4. The answer need not appear in a specific final-answer format.\par
5. Provide your reasoning.\par
6. Answer \textquotedbl yes\textquotedbl{} if the report contains the correct answer, and \textquotedbl no\textquotedbl{} if it
does not or contradicts it.\par\medskip
Output format:\par
correct: [yes/no]\par
reasoning: [your explanation]
\\
\bottomrule
\end{tabular}
\endgroup
\end{table}

\subsection{Judge Stability and Arbitration}
The shared binary evaluation implementation additionally supports a
three-vote stability audit. When this option is enabled, the primary judge
evaluates an item three times and its label is accepted only under unanimous
agreement. A split vote marks the item as disputed and invokes an independent
arbiter once, using the same benchmark prompt and parser; the arbiter's label
becomes the final decision. This audit records all primary votes and whether
arbitration was used. It is distinct from the single-call protocol used for the
main results and was not used to produce those scores.

\flushbottom
\section{Qualitative Report Examples}
\label{app:examples}

This appendix shows what MIRA delivers for a user request and how the two
report modes differ. Appendix~\ref{app:examples-modes} runs one request, on
Roman maritime concrete, in both \(\mathtt{CONCISE}\) and \(\mathtt{DETAIL}\)
modes and compares the two runs: how each one researched the request, how
much of the request each report covers, and how deeply each treats the same
topic. Appendix~\ref{app:examples-reports} reproduces both reports in full,
with annotations that check every reference and every image against what the
run actually opened. Appendix~\ref{app:examples-cam} does the same for a
second request, on how CAM (crassulacean acid metabolism) photosynthesis evolved in desert plants: both
reports in full, annotated the same way. The comparison figures in
Appendix~\ref{app:examples-modes} cover the first request only. Each pair of
runs shares the harness and model, differs only in the selected contract,
and is separate from the quantitative runs.

\input{sections/showcase/style}

\subsection{How the Concise Mode and the Detail Mode Differ}
\label{app:examples-modes}

\paragraph{Tool-call trajectory analysis.}
Figure~\ref{fig:timeline-concrete} lays out both trajectories turn by turn
and shows two things. First, the selected contract, not the request, sets
how much research a run does. Both runs follow the same order (searches,
then page opening, then image opening, then a first submission at turn 5),
but the Detail-mode run made 69 calls over 12 turns against 35 over 8, and
most of the difference is image opening (39 against 10): the Detail
contract asks for at least ten images and eight sources, the Concise
contract for two of each. Second, the report gate changes what the run does
next instead of only judging the result. Both first drafts were rejected,
and each run went back to research aimed at what the gate flagged. The
Concise-mode draft cited three references the gate could not reach; the run
searched again, opened three new pages, and was accepted at turn 8. The
Detail-mode draft cited seven sources where its contract requires eight,
four of them unreachable; the run searched again, opened four pages and seven images, and
rewrote four sections around them, was rejected again at turn 9 for seven
sources, and was accepted at turn 12 after one more search and page open.

\raggedbottom
\begin{figure}[htbp]
\centering
\begin{tikzpicture}[font=\scriptsize\sffamily, x=1cm, y=1cm]
\node[anchor=west, inner sep=1pt] (lgsearch) at (0,0.40) {\tikz\fill[tlsearch, rounded corners=0.6pt] (0,0) rectangle (0.22,0.22);\,search};
\node[anchor=west, inner sep=1pt] (lgpage) at ([xshift=9pt]lgsearch.east) {\tikz\fill[tlpage, rounded corners=0.6pt] (0,0) rectangle (0.22,0.22);\,open (web page)};
\node[anchor=west, inner sep=1pt] (lgimage) at ([xshift=9pt]lgpage.east) {\tikz\fill[tlimage, rounded corners=0.6pt] (0,0) rectangle (0.22,0.22);\,open (image)};
\node[anchor=west, inner sep=1pt] (lgsubmit) at ([xshift=9pt]lgimage.east) {\tikz\fill[tlsubmit, rounded corners=0.6pt] (0,0) rectangle (0.22,0.22);\,submit\_report};
\draw[decorate, decoration={brace, amplitude=3pt}, black!60, line width=0.5pt] ([xshift=3pt]lgsubmit.east |- 0,0.78) -- ([xshift=3pt]lgsubmit.east |- 0,0.02);
\node[anchor=west, inner sep=1pt] at ([xshift=9pt]lgsubmit.east |- 0,0.60) {\tikz{\fill[tlsubmit, rounded corners=0.6pt] (0,0) rectangle (0.22,0.22);\draw[white, line width=0.8pt, line cap=round] (0.059,0.059) -- (0.161,0.161) (0.059,0.161) -- (0.161,0.059);}\,rejected by the report gate};
\node[anchor=west, inner sep=1pt] at ([xshift=9pt]lgsubmit.east |- 0,0.20) {\tikz{\fill[tlsubmit, rounded corners=0.6pt] (0,0) rectangle (0.22,0.22);\draw[white, line width=0.8pt, line cap=round, line join=round] (0.048,0.114) -- (0.092,0.066) -- (0.172,0.158);}\,accepted};
\fill[tlband, rounded corners=1.5pt] (4.368,-2.150) rectangle (6.899,-0.480);
\node[anchor=south west, inner sep=1pt, font=\tiny\sffamily, text=black!60] at (4.368,-0.480) {revision after the gate rejects a draft};
\node[anchor=east, font=\scriptsize\sffamily\bfseries, text=rframe] at (-0.15,-1.40) {CONCISE};
\draw[black!25, line width=0.4pt] (0,-2.17) -- (6.93,-2.17);
\fill[tlsearch, rounded corners=0.5pt] (0.173,-2.100) rectangle ++(0.240,0.240);
\fill[tlsearch, rounded corners=0.5pt] (0.453,-2.100) rectangle ++(0.240,0.240);
\fill[tlsearch, rounded corners=0.5pt] (0.173,-1.820) rectangle ++(0.240,0.240);
\fill[tlsearch, rounded corners=0.5pt] (0.453,-1.820) rectangle ++(0.240,0.240);
\node[anchor=north, inner sep=1pt, font=\tiny\sffamily, text=black!55] at (0.433,-2.190) {1};
\fill[tlpage, rounded corners=0.5pt] (1.040,-2.100) rectangle ++(0.240,0.240);
\fill[tlpage, rounded corners=0.5pt] (1.320,-2.100) rectangle ++(0.240,0.240);
\fill[tlpage, rounded corners=0.5pt] (1.040,-1.820) rectangle ++(0.240,0.240);
\fill[tlpage, rounded corners=0.5pt] (1.320,-1.820) rectangle ++(0.240,0.240);
\fill[tlpage, rounded corners=0.5pt] (1.040,-1.540) rectangle ++(0.240,0.240);
\fill[tlpage, rounded corners=0.5pt] (1.320,-1.540) rectangle ++(0.240,0.240);
\fill[tlpage, rounded corners=0.5pt] (1.040,-1.260) rectangle ++(0.240,0.240);
\node[anchor=north, inner sep=1pt, font=\tiny\sffamily, text=black!55] at (1.300,-2.190) {2};
\fill[tlimage, rounded corners=0.5pt] (1.907,-2.100) rectangle ++(0.240,0.240);
\fill[tlimage, rounded corners=0.5pt] (2.187,-2.100) rectangle ++(0.240,0.240);
\fill[tlimage, rounded corners=0.5pt] (1.907,-1.820) rectangle ++(0.240,0.240);
\fill[tlimage, rounded corners=0.5pt] (2.187,-1.820) rectangle ++(0.240,0.240);
\fill[tlimage, rounded corners=0.5pt] (1.907,-1.540) rectangle ++(0.240,0.240);
\fill[tlimage, rounded corners=0.5pt] (2.187,-1.540) rectangle ++(0.240,0.240);
\fill[tlimage, rounded corners=0.5pt] (1.907,-1.260) rectangle ++(0.240,0.240);
\fill[tlimage, rounded corners=0.5pt] (2.187,-1.260) rectangle ++(0.240,0.240);
\fill[tlimage, rounded corners=0.5pt] (1.907,-0.980) rectangle ++(0.240,0.240);
\fill[tlimage, rounded corners=0.5pt] (2.187,-0.980) rectangle ++(0.240,0.240);
\node[anchor=north, inner sep=1pt, font=\tiny\sffamily, text=black!55] at (2.167,-2.190) {3};
\fill[tlpage, rounded corners=0.5pt] (2.773,-2.100) rectangle ++(0.240,0.240);
\fill[tlpage, rounded corners=0.5pt] (3.053,-2.100) rectangle ++(0.240,0.240);
\fill[tlpage, rounded corners=0.5pt] (2.773,-1.820) rectangle ++(0.240,0.240);
\fill[tlpage, rounded corners=0.5pt] (3.053,-1.820) rectangle ++(0.240,0.240);
\fill[tlpage, rounded corners=0.5pt] (2.773,-1.540) rectangle ++(0.240,0.240);
\fill[tlpage, rounded corners=0.5pt] (3.053,-1.540) rectangle ++(0.240,0.240);
\node[anchor=north, inner sep=1pt, font=\tiny\sffamily, text=black!55] at (3.033,-2.190) {4};
\fill[tlsubmit, rounded corners=0.5pt] (3.780,-2.100) rectangle ++(0.240,0.240);
\draw[white, line width=0.8pt, line cap=round] (3.845,-2.035) -- (3.955,-1.925) (3.845,-1.925) -- (3.955,-2.035);
\node[anchor=north, inner sep=1pt, font=\tiny\sffamily, text=black!55] at (3.900,-2.190) {5};
\fill[tlsearch, rounded corners=0.5pt] (4.507,-2.100) rectangle ++(0.240,0.240);
\fill[tlsearch, rounded corners=0.5pt] (4.787,-2.100) rectangle ++(0.240,0.240);
\fill[tlsearch, rounded corners=0.5pt] (4.507,-1.820) rectangle ++(0.240,0.240);
\node[anchor=north, inner sep=1pt, font=\tiny\sffamily, text=black!55] at (4.767,-2.190) {6};
\fill[tlpage, rounded corners=0.5pt] (5.373,-2.100) rectangle ++(0.240,0.240);
\fill[tlpage, rounded corners=0.5pt] (5.653,-2.100) rectangle ++(0.240,0.240);
\fill[tlpage, rounded corners=0.5pt] (5.373,-1.820) rectangle ++(0.240,0.240);
\node[anchor=north, inner sep=1pt, font=\tiny\sffamily, text=black!55] at (5.633,-2.190) {7};
\fill[tlsubmit, rounded corners=0.5pt] (6.380,-2.100) rectangle ++(0.240,0.240);
\draw[white, line width=0.8pt, line cap=round, line join=round] (6.433,-1.975) -- (6.481,-2.028) -- (6.567,-1.927);
\node[anchor=north, inner sep=1pt, font=\tiny\sffamily, text=black!55] at (6.500,-2.190) {8};
\node[anchor=north, inner sep=1pt, align=center, font=\tiny\sffamily, text=tlsubmit] (gCONCISE5) at (3.900,-2.520) {gate: 3 references unreachable};
\fill[tlsearch] (11.000,-1.560) rectangle ++(0.680,0.32);
\node[text=black!75, font=\tiny\sffamily\bfseries, inner sep=0pt] at (11.340,-1.400) {7};
\fill[tlpage] (11.680,-1.560) rectangle ++(1.554,0.32);
\node[text=black!75, font=\tiny\sffamily\bfseries, inner sep=0pt] at (12.457,-1.400) {16};
\fill[tlimage] (13.234,-1.560) rectangle ++(0.971,0.32);
\node[text=white, font=\tiny\sffamily\bfseries, inner sep=0pt] at (13.720,-1.400) {10};
\fill[tlsubmit] (14.206,-1.560) rectangle ++(0.194,0.32);
\node[anchor=west, font=\scriptsize\sffamily] at (11.000,-1.040) {35 calls, 8 turns};
\fill[tlband, rounded corners=1.5pt] (4.368,-6.390) rectangle (10.365,-3.320);
\node[anchor=south west, inner sep=1pt, font=\tiny\sffamily, text=black!60] at (4.368,-3.320) {revision after the gate rejects a draft};
\node[anchor=east, font=\scriptsize\sffamily\bfseries, text=rframe] at (-0.15,-4.94) {DETAIL};
\draw[black!25, line width=0.4pt] (0,-6.41) -- (10.40,-6.41);
\fill[tlsearch, rounded corners=0.5pt] (0.173,-6.340) rectangle ++(0.240,0.240);
\fill[tlsearch, rounded corners=0.5pt] (0.453,-6.340) rectangle ++(0.240,0.240);
\fill[tlsearch, rounded corners=0.5pt] (0.173,-6.060) rectangle ++(0.240,0.240);
\fill[tlsearch, rounded corners=0.5pt] (0.453,-6.060) rectangle ++(0.240,0.240);
\fill[tlsearch, rounded corners=0.5pt] (0.173,-5.780) rectangle ++(0.240,0.240);
\node[anchor=north, inner sep=1pt, font=\tiny\sffamily, text=black!55] at (0.433,-6.430) {1};
\fill[tlpage, rounded corners=0.5pt] (1.040,-6.340) rectangle ++(0.240,0.240);
\fill[tlpage, rounded corners=0.5pt] (1.320,-6.340) rectangle ++(0.240,0.240);
\fill[tlpage, rounded corners=0.5pt] (1.040,-6.060) rectangle ++(0.240,0.240);
\fill[tlpage, rounded corners=0.5pt] (1.320,-6.060) rectangle ++(0.240,0.240);
\fill[tlpage, rounded corners=0.5pt] (1.040,-5.780) rectangle ++(0.240,0.240);
\fill[tlpage, rounded corners=0.5pt] (1.320,-5.780) rectangle ++(0.240,0.240);
\fill[tlpage, rounded corners=0.5pt] (1.040,-5.500) rectangle ++(0.240,0.240);
\fill[tlpage, rounded corners=0.5pt] (1.320,-5.500) rectangle ++(0.240,0.240);
\fill[tlpage, rounded corners=0.5pt] (1.040,-5.220) rectangle ++(0.240,0.240);
\fill[tlpage, rounded corners=0.5pt] (1.320,-5.220) rectangle ++(0.240,0.240);
\node[anchor=north, inner sep=1pt, font=\tiny\sffamily, text=black!55] at (1.300,-6.430) {2};
\fill[tlsearch, rounded corners=0.5pt] (1.907,-6.340) rectangle ++(0.240,0.240);
\fill[tlsearch, rounded corners=0.5pt] (2.187,-6.340) rectangle ++(0.240,0.240);
\fill[tlsearch, rounded corners=0.5pt] (1.907,-6.060) rectangle ++(0.240,0.240);
\fill[tlimage, rounded corners=0.5pt] (2.187,-6.060) rectangle ++(0.240,0.240);
\fill[tlimage, rounded corners=0.5pt] (1.907,-5.780) rectangle ++(0.240,0.240);
\fill[tlimage, rounded corners=0.5pt] (2.187,-5.780) rectangle ++(0.240,0.240);
\fill[tlimage, rounded corners=0.5pt] (1.907,-5.500) rectangle ++(0.240,0.240);
\fill[tlimage, rounded corners=0.5pt] (2.187,-5.500) rectangle ++(0.240,0.240);
\fill[tlimage, rounded corners=0.5pt] (1.907,-5.220) rectangle ++(0.240,0.240);
\fill[tlimage, rounded corners=0.5pt] (2.187,-5.220) rectangle ++(0.240,0.240);
\fill[tlimage, rounded corners=0.5pt] (1.907,-4.940) rectangle ++(0.240,0.240);
\fill[tlimage, rounded corners=0.5pt] (2.187,-4.940) rectangle ++(0.240,0.240);
\fill[tlimage, rounded corners=0.5pt] (1.907,-4.660) rectangle ++(0.240,0.240);
\fill[tlimage, rounded corners=0.5pt] (2.187,-4.660) rectangle ++(0.240,0.240);
\fill[tlimage, rounded corners=0.5pt] (1.907,-4.380) rectangle ++(0.240,0.240);
\fill[tlimage, rounded corners=0.5pt] (2.187,-4.380) rectangle ++(0.240,0.240);
\fill[tlimage, rounded corners=0.5pt] (1.907,-4.100) rectangle ++(0.240,0.240);
\fill[tlimage, rounded corners=0.5pt] (2.187,-4.100) rectangle ++(0.240,0.240);
\fill[tlimage, rounded corners=0.5pt] (1.907,-3.820) rectangle ++(0.240,0.240);
\fill[tlimage, rounded corners=0.5pt] (2.187,-3.820) rectangle ++(0.240,0.240);
\node[anchor=north, inner sep=1pt, font=\tiny\sffamily, text=black!55] at (2.167,-6.430) {3};
\fill[tlimage, rounded corners=0.5pt] (2.773,-6.340) rectangle ++(0.240,0.240);
\fill[tlimage, rounded corners=0.5pt] (3.053,-6.340) rectangle ++(0.240,0.240);
\fill[tlimage, rounded corners=0.5pt] (2.773,-6.060) rectangle ++(0.240,0.240);
\fill[tlimage, rounded corners=0.5pt] (3.053,-6.060) rectangle ++(0.240,0.240);
\fill[tlimage, rounded corners=0.5pt] (2.773,-5.780) rectangle ++(0.240,0.240);
\fill[tlimage, rounded corners=0.5pt] (3.053,-5.780) rectangle ++(0.240,0.240);
\fill[tlimage, rounded corners=0.5pt] (2.773,-5.500) rectangle ++(0.240,0.240);
\fill[tlimage, rounded corners=0.5pt] (3.053,-5.500) rectangle ++(0.240,0.240);
\fill[tlimage, rounded corners=0.5pt] (2.773,-5.220) rectangle ++(0.240,0.240);
\fill[tlimage, rounded corners=0.5pt] (3.053,-5.220) rectangle ++(0.240,0.240);
\fill[tlimage, rounded corners=0.5pt] (2.773,-4.940) rectangle ++(0.240,0.240);
\fill[tlimage, rounded corners=0.5pt] (3.053,-4.940) rectangle ++(0.240,0.240);
\fill[tlimage, rounded corners=0.5pt] (2.773,-4.660) rectangle ++(0.240,0.240);
\fill[tlimage, rounded corners=0.5pt] (3.053,-4.660) rectangle ++(0.240,0.240);
\fill[tlimage, rounded corners=0.5pt] (2.773,-4.380) rectangle ++(0.240,0.240);
\node[anchor=north, inner sep=1pt, font=\tiny\sffamily, text=black!55] at (3.033,-6.430) {4};
\fill[tlsubmit, rounded corners=0.5pt] (3.780,-6.340) rectangle ++(0.240,0.240);
\draw[white, line width=0.8pt, line cap=round] (3.845,-6.275) -- (3.955,-6.165) (3.845,-6.165) -- (3.955,-6.275);
\node[anchor=north, inner sep=1pt, font=\tiny\sffamily, text=black!55] at (3.900,-6.430) {5};
\fill[tlsearch, rounded corners=0.5pt] (4.507,-6.340) rectangle ++(0.240,0.240);
\fill[tlsearch, rounded corners=0.5pt] (4.787,-6.340) rectangle ++(0.240,0.240);
\fill[tlsearch, rounded corners=0.5pt] (4.507,-6.060) rectangle ++(0.240,0.240);
\node[anchor=north, inner sep=1pt, font=\tiny\sffamily, text=black!55] at (4.767,-6.430) {6};
\fill[tlpage, rounded corners=0.5pt] (5.373,-6.340) rectangle ++(0.240,0.240);
\fill[tlpage, rounded corners=0.5pt] (5.653,-6.340) rectangle ++(0.240,0.240);
\fill[tlpage, rounded corners=0.5pt] (5.373,-6.060) rectangle ++(0.240,0.240);
\fill[tlpage, rounded corners=0.5pt] (5.653,-6.060) rectangle ++(0.240,0.240);
\node[anchor=north, inner sep=1pt, font=\tiny\sffamily, text=black!55] at (5.633,-6.430) {7};
\fill[tlimage, rounded corners=0.5pt] (6.240,-6.340) rectangle ++(0.240,0.240);
\fill[tlimage, rounded corners=0.5pt] (6.520,-6.340) rectangle ++(0.240,0.240);
\fill[tlimage, rounded corners=0.5pt] (6.240,-6.060) rectangle ++(0.240,0.240);
\fill[tlimage, rounded corners=0.5pt] (6.520,-6.060) rectangle ++(0.240,0.240);
\fill[tlimage, rounded corners=0.5pt] (6.240,-5.780) rectangle ++(0.240,0.240);
\fill[tlimage, rounded corners=0.5pt] (6.520,-5.780) rectangle ++(0.240,0.240);
\fill[tlimage, rounded corners=0.5pt] (6.240,-5.500) rectangle ++(0.240,0.240);
\node[anchor=north, inner sep=1pt, font=\tiny\sffamily, text=black!55] at (6.500,-6.430) {8};
\fill[tlsubmit, rounded corners=0.5pt] (7.247,-6.340) rectangle ++(0.240,0.240);
\draw[white, line width=0.8pt, line cap=round] (7.311,-6.275) -- (7.422,-6.165) (7.311,-6.165) -- (7.422,-6.275);
\node[anchor=north, inner sep=1pt, font=\tiny\sffamily, text=black!55] at (7.367,-6.430) {9};
\fill[tlsearch, rounded corners=0.5pt] (8.113,-6.340) rectangle ++(0.240,0.240);
\node[anchor=north, inner sep=1pt, font=\tiny\sffamily, text=black!55] at (8.233,-6.430) {10};
\fill[tlpage, rounded corners=0.5pt] (8.980,-6.340) rectangle ++(0.240,0.240);
\node[anchor=north, inner sep=1pt, font=\tiny\sffamily, text=black!55] at (9.100,-6.430) {11};
\fill[tlsubmit, rounded corners=0.5pt] (9.847,-6.340) rectangle ++(0.240,0.240);
\draw[white, line width=0.8pt, line cap=round, line join=round] (9.899,-6.215) -- (9.947,-6.268) -- (10.034,-6.167);
\node[anchor=north, inner sep=1pt, font=\tiny\sffamily, text=black!55] at (9.967,-6.430) {12};
\node[anchor=north, inner sep=1pt, align=center, font=\tiny\sffamily, text=tlsubmit] (gDETAIL5) at (3.900,-6.760) {gate: 7 of 8 required sources\\4 references unreachable};
\node[anchor=north, inner sep=1pt, align=center, font=\tiny\sffamily, text=tlsubmit] (gDETAIL9) at (7.367,-6.760) {gate: 7 of 8 required sources};
\fill[tlsearch] (11.000,-5.100) rectangle ++(0.591,0.32);
\node[text=black!75, font=\tiny\sffamily\bfseries, inner sep=0pt] at (11.296,-4.940) {12};
\fill[tlpage] (11.591,-5.100) rectangle ++(0.739,0.32);
\node[text=black!75, font=\tiny\sffamily\bfseries, inner sep=0pt] at (11.961,-4.940) {15};
\fill[tlimage] (12.330,-5.100) rectangle ++(1.922,0.32);
\node[text=white, font=\tiny\sffamily\bfseries, inner sep=0pt] at (13.291,-4.940) {39};
\fill[tlsubmit] (14.252,-5.100) rectangle ++(0.148,0.32);
\node[anchor=west, font=\scriptsize\sffamily] at (11.000,-4.580) {69 calls, 12 turns};
\node[anchor=north, font=\tiny\sffamily, text=black!55] at (5.20,-7.26) {assistant turn};
\end{tikzpicture}
\caption{\textbf{Tool calls for the Roman-concrete request.} Each column is one assistant turn, numbered below the axis; each square is one tool call made in that turn, two to a row, shaded by tool from light (search) to dark (\texttt{submit\_report}). A turn can issue several calls at once; turn 3 of the \(\mathtt{DETAIL}\) run, for example, issues 20. The \texttt{open} tool appears twice because it opens either a web page or an image found on an opened page. A cross on a \texttt{submit\_report} square means the report gate rejected that draft, and a tick means it accepted it; the gate's reason is printed under the axis, and the tinted band covers the turns spent revising after a rejection. The bar on the right gives each run's total calls by tool.}
\label{fig:timeline-concrete}
\end{figure}
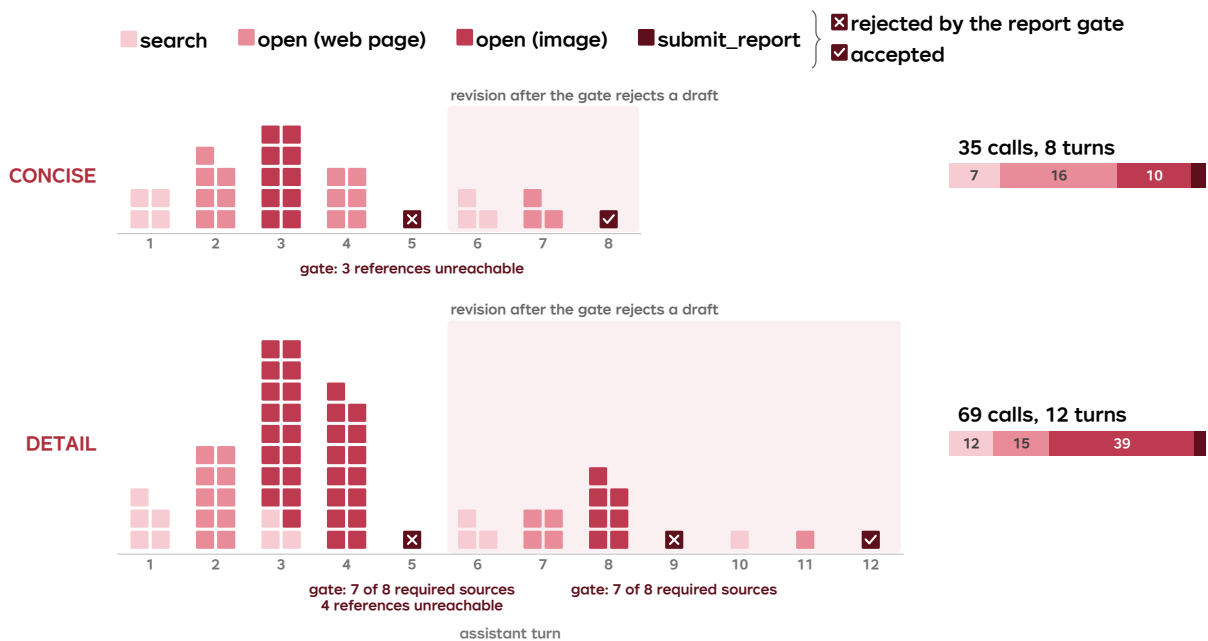

\paragraph{Detail covers the whole request.}
The two modes trade breadth for length: the Concise-mode report answers the
core of the request in a few sentences, while the Detail-mode report gives
every part of the request its own treatment. The request names eight
concepts, and Figure~\ref{fig:coverage-concrete} counts, for each of them,
the report sentences that mention it. The Detail-mode report mentions all
eight in 59 sentences. The Concise-mode report mentions seven in 10
sentences and leaves out the microstructural studies, keeping to the three
mechanisms the request asks about.

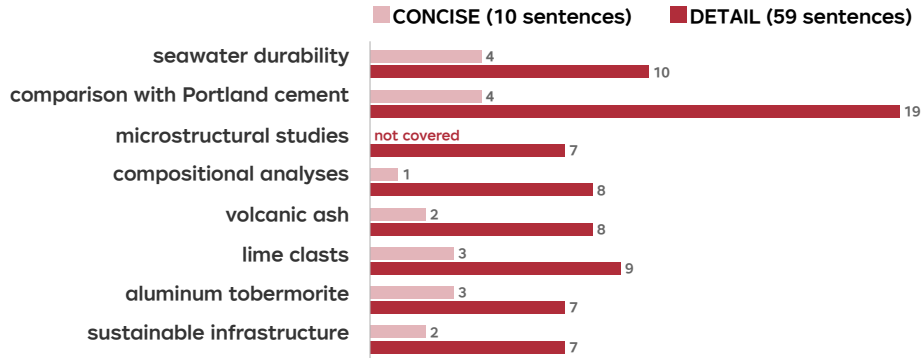
\begin{figure}[htbp]
\centering
\begin{tikzpicture}[font=\scriptsize\sffamily, x=1cm, y=1cm]
\node[anchor=west, inner sep=1pt] (lc) at (0,0.45) {\tikz\fill[rframe!35] (0,0) rectangle (0.22,0.22);\,CONCISE (10 sentences)};
\node[anchor=west, inner sep=1pt] at ([xshift=12pt]lc.east) {\tikz\fill[rframe] (0,0) rectangle (0.22,0.22);\,DETAIL (59 sentences)};
\node[anchor=east, text=black!80] at (-0.15,-0.09) {seawater durability};
\fill[rframe!35] (0,0.000) rectangle ++(1.474,-0.170);
\node[anchor=west, inner sep=1.5pt, font=\tiny\sffamily, text=black!60] at (1.474,-0.085) {4};
\fill[rframe] (0,-0.200) rectangle ++(3.684,-0.170);
\node[anchor=west, inner sep=1.5pt, font=\tiny\sffamily, text=black!60] at (3.684,-0.285) {10};
\node[anchor=east, text=black!80] at (-0.15,-0.60) {comparison with Portland cement};
\fill[rframe!35] (0,-0.520) rectangle ++(1.474,-0.170);
\node[anchor=west, inner sep=1.5pt, font=\tiny\sffamily, text=black!60] at (1.474,-0.605) {4};
\fill[rframe] (0,-0.720) rectangle ++(7.000,-0.170);
\node[anchor=west, inner sep=1.5pt, font=\tiny\sffamily, text=black!60] at (7.000,-0.805) {19};
\node[anchor=east, text=black!80] at (-0.15,-1.12) {microstructural studies};
\node[anchor=west, inner sep=1.5pt, font=\tiny\sffamily, text=rframe] at (0.000,-1.125) {not covered};
\fill[rframe] (0,-1.240) rectangle ++(2.579,-0.170);
\node[anchor=west, inner sep=1.5pt, font=\tiny\sffamily, text=black!60] at (2.579,-1.325) {7};
\node[anchor=east, text=black!80] at (-0.15,-1.65) {compositional analyses};
\fill[rframe!35] (0,-1.560) rectangle ++(0.368,-0.170);
\node[anchor=west, inner sep=1.5pt, font=\tiny\sffamily, text=black!60] at (0.368,-1.645) {1};
\fill[rframe] (0,-1.760) rectangle ++(2.947,-0.170);
\node[anchor=west, inner sep=1.5pt, font=\tiny\sffamily, text=black!60] at (2.947,-1.845) {8};
\node[anchor=east, text=black!80] at (-0.15,-2.17) {volcanic ash};
\fill[rframe!35] (0,-2.080) rectangle ++(0.737,-0.170);
\node[anchor=west, inner sep=1.5pt, font=\tiny\sffamily, text=black!60] at (0.737,-2.165) {2};
\fill[rframe] (0,-2.280) rectangle ++(2.947,-0.170);
\node[anchor=west, inner sep=1.5pt, font=\tiny\sffamily, text=black!60] at (2.947,-2.365) {8};
\node[anchor=east, text=black!80] at (-0.15,-2.69) {lime clasts};
\fill[rframe!35] (0,-2.600) rectangle ++(1.105,-0.170);
\node[anchor=west, inner sep=1.5pt, font=\tiny\sffamily, text=black!60] at (1.105,-2.685) {3};
\fill[rframe] (0,-2.800) rectangle ++(3.316,-0.170);
\node[anchor=west, inner sep=1.5pt, font=\tiny\sffamily, text=black!60] at (3.316,-2.885) {9};
\node[anchor=east, text=black!80] at (-0.15,-3.21) {aluminum tobermorite};
\fill[rframe!35] (0,-3.120) rectangle ++(1.105,-0.170);
\node[anchor=west, inner sep=1.5pt, font=\tiny\sffamily, text=black!60] at (1.105,-3.205) {3};
\fill[rframe] (0,-3.320) rectangle ++(2.579,-0.170);
\node[anchor=west, inner sep=1.5pt, font=\tiny\sffamily, text=black!60] at (2.579,-3.405) {7};
\node[anchor=east, text=black!80] at (-0.15,-3.73) {sustainable infrastructure};
\fill[rframe!35] (0,-3.640) rectangle ++(0.737,-0.170);
\node[anchor=west, inner sep=1.5pt, font=\tiny\sffamily, text=black!60] at (0.737,-3.725) {2};
\fill[rframe] (0,-3.840) rectangle ++(2.579,-0.170);
\node[anchor=west, inner sep=1.5pt, font=\tiny\sffamily, text=black!60] at (2.579,-3.925) {7};
\draw[black!30, line width=0.4pt] (0,0.12) -- (0,-4.08);
\end{tikzpicture}
\caption{\textbf{Coverage of the request.} For each of the eight concepts the request names, the number of report sentences that mention it, counted by keyword match over the report body (headings, image captions, and the reference list excluded). The \(\mathtt{CONCISE}\) report touches seven of the eight in 10 sentences; the \(\mathtt{DETAIL}\) report touches all eight in 59 sentences and is the only one to describe microstructural studies.}
\label{fig:coverage-concrete}
\end{figure}

\paragraph{Detail goes deeper on one topic.}
On a topic both reports cover, the two modes agree on the core explanation;
the Detail mode adds context and mechanism, not different facts.
Figure~\ref{fig:compare-concrete} collects everything each report says about
lime clasts. Both explain how hot mixing with quicklime creates the clasts
and how they heal cracks. The Detail-mode report adds the earlier
reading of the clasts as poor-quality raw material, their role as a
long-term calcium source for pore filling, and two further figures from the
same MIT page.

\begin{figure}[htbp]
\centering
\footnotesize
\setlength{\tabcolsep}{5pt}
\renewcommand{\arraystretch}{1.15}
\begin{tabular}{@{}p{0.13\linewidth}p{0.36\linewidth}p{0.45\linewidth}@{}}
\toprule
& {\sffamily\bfseries\color{rframe}CONCISE}
& {\sffamily\bfseries\color{rframe}DETAIL} \\ \midrule
\cmprow{How they form}
{Quicklime used ``rather than pre-slaked lime'' creates them~[1].}
{``Hot mixing'' with quicklime (CaO) \dadd{rather than only slaked lime; the exothermic reaction} creates them~[2].}
\cmprow{What they are}
{``bright white, highly reactive calcium carbonate inclusions''~[1].}
{``millimeter-scale bright white mineral features''\dadd{, macroscopic sources of reactive calcium}~[2].}
\cmprow{Earlier view}
{Clasts ``were once dismissed as sloppy mixing.''}
{Dismissed as sloppy mixing \dadd{or poor-quality raw materials, now seen as purposefully incorporated}~[2].}
\cmprow{How they heal cracks}
{Water dissolves calcium from the clasts, which ``recrystallizes to seal the damage''~[2].}
{Seawater mobilizes calcium from the clasts, \dadd{and reactive hydration heals the cracks and remodels the interfacial zones; the clasts also feed long-term pore filling and post-pozzolanic reactivity}~[2].}
\cmprow{Modern use}
{Lime-clast modern concretes ``demonstrate self-healing potential''~[2].}
{Modern lime-clast mixtures show self-healing potential~[2], \dadd{framed as an ``antiqua-inspired'' approach to new materials}.}
\end{tabular}
\caption{\textbf{One topic in both modes: lime clasts.} Everything each report
says about lime clasts, aligned point by point. Quotation marks mark text
copied from the report; other cells are condensed from it without new
content. \protect\dadd{Pink} marks what only the \(\mathtt{DETAIL}\) report
says. Reference numbers follow each report's own list.}
\label{fig:compare-concrete}
\end{figure}

\FloatBarrier
\subsection{Concise and Detail Reports for the Roman Concrete Request}
\label{app:examples-reports}

Both reports are reproduced in full, here and in
Appendix~\ref{app:examples-cam}, typeset as text from the delivered
Markdown with internal paths omitted; the report text itself is unchanged.
Each report opens with a header card that gives its mode, its delivered
statistics, and the user request, and closes with a gate record that lists
every submission and its result. The annotations added for this appendix are
explained below.

\begin{rlegendbox}
\clue{Marker}: a concept named in the request, at its first appearance.
\fact{Wavy line}: a specific claim whose source page was reopened for this appendix.
\textbf{TRACE}: when the source page and the image were opened, read from the
trajectory, and why the image was used.
\textbf{NOTE}: our commentary.
{\color{rframe}\textbf{Red boxes}}: the part of an image the text relies on.
{\sffamily\bfseries\color{rframe}[Image~\(n\)]}: placed after the sentence that image \(n\) supports.
The delivered Markdown places images inline without numbers; the image
numbers, these markers, and the per-image source list at the end of each
report are added for this appendix.
\textbf{REPORT GATE RECORD}: each \texttt{submit\_report} call and its checks.
\par\vspace{2pt}
\textbf{Reference check.} Every reference URL in the reports of this
appendix was reopened and resolves to the cited page; for
Appendix~\ref{app:examples-cam}, whose journal pages block automated access,
claims were checked against the page text and captions captured during the
run. The report gate checks that references are
reachable and that their number meets the contract; it does not check that a
page supports the sentence that cites it. The NOTE boxes mark each claim that
is missing from, or only partly on, its cited page, and each image whose
description does not match its pixels.
The source count in the header card and the gate record counts distinct
source pages, including image source pages that the reference list does not
cite.
\end{rlegendbox}

\begin{reportcard}{CONCISE}{1,722 chars \textperiodcentered{} 3 images \textperiodcentered{} 5 sources \textperiodcentered{} 2 submissions}{Why does Roman maritime concrete demonstrate superior long-term durability in seawater compared to many modern Portland cement formulations, and what do microstructural studies and compositional analyses reveal about the role of volcanic ash, lime clasts, and aluminum tobermorite formation? I need to compare these ancient and modern materials systems for a unit on sustainable infrastructure materials.}
\rtitle{Why Roman Maritime Concrete Outlasts Modern Portland Cement}
Roman maritime concrete survives millennia in \clue{seawater} because its chemistry actively strengthens over time, unlike modern \clue{Portland cement} that degrades. The secret lies in \clue{volcanic ash}, \clue{lime clasts}, and \clue{aluminum-tobermorite} formation.

\rsec{Volcanic Ash and Hot Mixing}
\rnote{Reference check. The crack-sealing mechanism at the end of this section (calcium dissolves and recrystallizes) is described on the MIT page [1]; the Scientific American page [2] reports only that the clasts enabled self-healing.}
Romans mixed lime with volcanic ash (pozzolana) and seawater, triggering an exothermic reaction. A 2023 MIT study revealed they \fact{deliberately used quicklime rather than pre-slaked lime}, creating ``lime clasts''---bright white, highly reactive calcium carbonate inclusions \rcite{1}.\rimgref{1} These clasts were once dismissed as sloppy mixing, but they provide a self-healing reservoir: when cracks form, water dissolves calcium from the clasts, which recrystallizes to seal the damage \rcite{2}.

\rtrace{turn 2 \texttt{open(url)} \(\rightarrow\) turn 3 \texttt{open(id=page\_img\_1)}}{Source page: \emph{Riddle solved: Why was Roman concrete so durable? | MIT News | Massachusetts Institute of Technology}; caption on the page: ``On left, Roman concrete structures of Privernum, Italy with icy puddles, orange sky, and rural location. On right, the concrete fragment is\ldots''\par\textbf{Why it was used.} The one image that shows a lime clast: the red region of the color-mapped fragment.}
\rimagebox{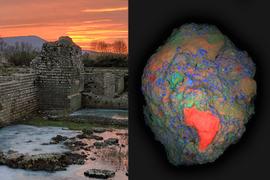}{Roman concrete structures at Privernum, Italy, with a colorized fragment showing calcium-rich regions in red}{\rbox{0.67}{0.17}{0.83}{0.42}{}}
\rsec{Al-Tobermorite: The Seawater Crystal}
\rnote{Reference check. The gate rejected the first draft because three references were unreachable. The run then opened three new pages and put them in the reference list, but kept the text and the citation numbers. The crystal-growth claim below is on the ASME page [4], not on [3], and the steel-corrosion clause cited to [4] is not on that page.}
When seawater percolates through Roman concrete, it triggers formation of \textbf{Al-tobermorite} and \textbf{phillipsite}---rare interlocking crystals that grow inside micro-cracks and voids \rcite{3}.\rimgref{2} These minerals bind the matrix tighter over centuries, making submerged sections stronger than air-exposed ones. Modern concrete lacks this regenerative mechanism; seawater instead corrodes steel and breaks down its binding matrix \rcite{4}.

\rtrace{turn 2 \texttt{open(url)} \(\rightarrow\) turn 3 \texttt{open(id=page\_img\_21)}}{Source page: \emph{Learning from Roman Seawater Concrete}; caption on the page: ``Drill core of mortar consisting of volcanic ash and hydrated lime, showing yellowish particles of pumice, dark gray fragments of lava, and w\ldots''\par\textbf{Why it was used.} Pairs the drill core with an SEM view of the Al-tobermorite crystals that the section is about.}
\rimagebox{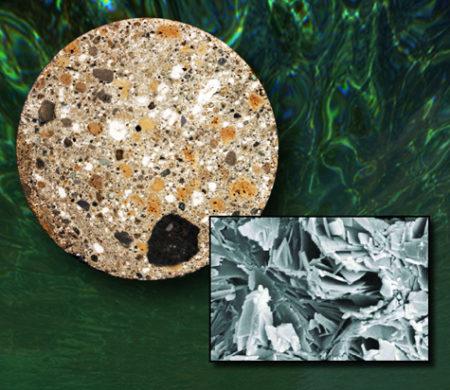}{Drill core of Roman mortar with volcanic ash, pumice, lava fragments, and relict lime, plus SEM image of Al-tobermorite crystals}{\rbox{0.46}{0.07}{0.96}{0.45}{}}
\rsec{Sustainable Infrastructure Implications}
\rnote{Reference check. Page [3] supports the 10\% lime and the lower temperature. The sentence cited to [2] is partly supported: Scientific American reports healed hot-mixed samples, but the role of Al-tobermorite as a model for low-carbon composites comes from the Berkeley Lab ALS page, which the run opened but did not cite.}
Roman concrete used \fact{\textasciitilde{}10\% lime by weight, produced at lower temperatures than Portland cement}, and leveraged locally available volcanic materials \rcite{3}. Modern research is now replicating these principles: lime clast-containing modern concretes demonstrate self-healing potential, and Al-tobermorite serves as a model for high-performance, low-carbon cementitious composites \rcite{2}.\rimgref{3}

\rtrace{turn 2 \texttt{open(url)} \(\rightarrow\) turn 3 \texttt{open(id=page\_img\_2)}}{Source page: \emph{Riddle solved: Why was Roman concrete so durable? | MIT News | Massachusetts Institute of Technology}; caption on the page: ``The concrete fragment is colorized with rainbow colors, including a prominent section colored red.''\par\textbf{Why it was used.} A crop of the fragment in Image~1. The red, calcium-rich region named in the caption lies outside this crop.}
\rimagebox{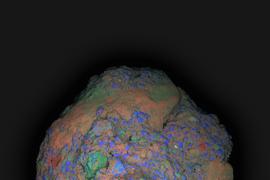}{Colorized concrete fragment with rainbow elemental mapping showing calcium-rich zones}{}
\rsec{References}
\begin{rlist}
\item \rcite{1} Riddle solved: Why was Roman concrete so durable? | MIT News\rurl{https://news.mit.edu/2023/roman-concrete-durability-lime-casts-0106}
\item \rcite{2} Ancient Roman Concrete Has 'Self-Healing' Capabilities | Scientific American\rurl{https://www.scientificamerican.com/article/ancient-roman-concrete-has-self-healing-capabilities/}
\item \rcite{3} Roman Seawater Concrete Holds the Secret to Cutting Carbon Emissions - Berkeley Lab\rurl{https://newscenter.lbl.gov/2013/06/04/roman-concrete/}
\item \rcite{4} Ancient Roman Concrete Stands the Test of Time - ASME\rurl{https://www.asme.org/topics-resources/content/ancient-roman-concrete-stands-test-time}
\end{rlist}
\rimgsources{\rimgsrc{Images~1, 3}{Riddle solved: Why was Roman concrete so durable? | MIT News | Massachusetts Institute of Technology}{https://news.mit.edu/2023/roman-concrete-durability-lime-casts-0106}
\rimgsrc{Image~2}{Learning from Roman Seawater Concrete}{https://als.lbl.gov/learning-from-roman-concrete/}}
\rgate{\item Submission 1 (turn 5): {\color{rframe}\textbf{WARN}} --- 1,698 chars, 3~images, 5~sources; \texttt{reference\_unreachable}\,\(\times\)3
\item Submission 2 (turn 8): {\color{tok}\textbf{OK}} --- 1,722 chars, 3~images, 5~sources}

\end{reportcard}

\begin{reportcard}{DETAIL}{8,063 chars \textperiodcentered{} 13 images \textperiodcentered{} 8 sources \textperiodcentered{} 3 submissions}{Why does Roman maritime concrete demonstrate superior long-term durability in seawater compared to many modern Portland cement formulations, and what do microstructural studies and compositional analyses reveal about the role of volcanic ash, lime clasts, and aluminum tobermorite formation? I need to compare these ancient and modern materials systems for a unit on sustainable infrastructure materials.}
\rtitle{Roman Maritime Concrete: Why Ancient Seawater Structures Outlast Modern Portland Cement}
\rsec{Introduction}
Roman maritime concrete has survived for over two millennia in aggressive \clue{seawater} environments, while many modern \clue{Portland cement} structures deteriorate within decades. This report examines the material science behind Roman concrete's exceptional durability, focusing on the roles of \clue{volcanic ash}, \clue{lime clasts}, and \clue{aluminum tobermorite} formation revealed by modern \clue{microstructural} and \clue{compositional analyses}. Understanding these ancient formulations offers critical insights for developing \clue{sustainable}, long-lasting infrastructure materials.

\rsec{Composition and Microstructure: The Roman Recipe}
Roman concrete, or \emph{opus caementicium}, differs fundamentally from modern Portland cement in its composition and manufacturing process. While modern concrete relies on Portland cement clinker---produced by calcining limestone and clay at \fact{approximately 1450\textdegree{}C}---Roman concrete used lime and volcanic ash (pozzolana) mixed with water and coarse aggregates \rcite{1}. The Romans employed a ``hot mixing'' technique, incorporating quicklime (CaO) directly into the mixture rather than using only slaked lime \rcite{2}. This exothermic reaction generated temperatures sufficient to create distinctive millimeter-scale bright white mineral features known as ``lime clasts'' \rcite{2}.

\rtrace{turn 2 \texttt{open(url)} \(\rightarrow\) turn 3 \texttt{open(id=page\_img\_7)}}{Source page: \emph{Learning from Roman Seawater Concrete}; caption on the page: ``Drill core of mortar consisting of volcanic ash and hydrated lime, showing yellowish particles of pumice, dark gray fragments of lava, and w\ldots''\par\textbf{Why it was used.} Pairs the drill core with an SEM view of Al-tobermorite, matching the composite structure the paragraph describes.}
\rimagebox{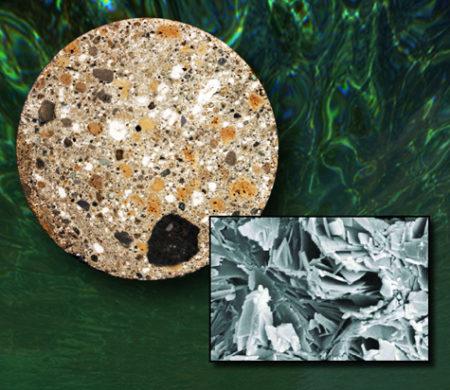}{Drill core of Roman mortar showing volcanic ash, pumice, lava fragments, and relict lime particles, with inset SEM image of Al-tobermorite crystals}{\rbox{0.46}{0.07}{0.96}{0.45}{}}
The drill core shown above illustrates the heterogeneous microstructure of Roman maritime concrete from Pozzuoli Bay.\rimgref{1} The yellowish pumice particles, dark gray lava fragments, and white relict lime particles are all bound by a light gray cementitious matrix \rcite{1}. This composite structure differs dramatically from the relatively homogeneous binder-aggregate system of modern concrete.

\rtrace{turn 7 \texttt{open(url)} \(\rightarrow\) turn 8 \texttt{open(id=page\_img\_46)}}{Source page: \emph{Ancient Roman Concrete Stands the Test of Time - ASME}; caption on the page: ``An ancient pier is still standing in a bay in Italy, and researchers have studied samples of the concrete to explore the secrets of its long\ldots''\par\textbf{Why it was used.} Shows the kind of harbor structure that was sampled; it sets the scene rather than carrying a measurement. It shows the same pier and drilling rig as Image~6, photographed from a different position.}
\rimagebox{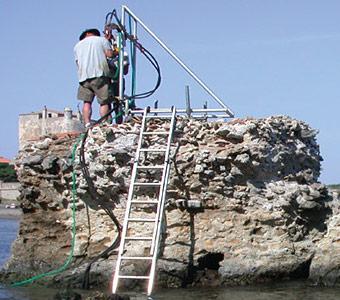}{Ancient pier still standing in a bay in Italy, with researchers studying samples of the concrete}{}
Researchers from Lawrence Berkeley National Laboratory and the University of Utah analyzed samples of Roman concrete from harbor structures in Italy at Berkeley Lab's Advanced Light Source \rcite{5}.\rimgref{2} The ancient pier shown above demonstrates the remarkable longevity of these structures in marine environments.

\rsec{The Role of Volcanic Ash: Pozzolanic Reactivity}
\rnote{Reference check. Silica and alumina are on the Wisconsin page [3]; the Pozzuoli Bay location comes from [1]. The pozzolanic-reaction sentence is only partly on [7], which says that seawater forms aluminous tobermorite and phillipsite.}
Volcanic ash, particularly from the Pozzuoli Bay region, contains both silica and alumina that proved crucial for the chemical reactions responsible for Roman concrete's durability \rcite{3}. When seawater contacts the concrete, a pozzolanic reaction occurs between the volcanic ash, lime, and saltwater to produce enduring cementitious hydrates \rcite{7}.

\rtrace{turn 7 \texttt{open(url)} \(\rightarrow\) turn 8 \texttt{open(id=page\_img\_67)}}{Source page: \emph{Why is Roman concrete more durable than modern concrete?}; caption on the page: ``Figure 1: Erosion due to sea water on concrete pylons. Image: Brian Robinson.''\par\textbf{Why it was used.} The report's only image of modern concrete, eroded by seawater.}
\rimagebox{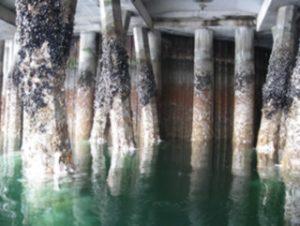}{Erosion due to sea water on modern concrete pylons, showing the degradation that modern concrete suffers}{}
Modern concrete is porous and degrades in contact with seawater.\rimgref{3} Seawater can seep into its pores, and when dried out the salts crystallize. The crystallization pressure of the salts produces stresses that can result in cracks and spalls \rcite{7}. Other chemical processes such as sulphate attack, lime leaching, and alkali-aggregate expansion all degrade modern concrete.

\rtrace{turn 7 \texttt{open(url)} \(\rightarrow\) turn 8 \texttt{open(id=page\_img\_68)}}{Source page: \emph{Why is Roman concrete more durable than modern concrete?}; caption on the page: ``Figure 2: A 2000 year old Roman jetty. Image: Art853.''\par\textbf{Why it was used.} A Roman jetty, set against the eroded modern pylons of Image~3.}
\rimagebox{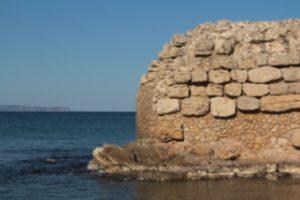}{A 2000 year old Roman jetty still standing, demonstrating the superior durability of Roman concrete}{}
In contrast, 2000-year-old concrete constructed during the Roman Empire is still going strong \rcite{7}.\rimgref{4} The Roman concrete samples were found to contain rare aluminous tobermorite and phillipsite crystals that strengthen the material over time.

\rsec{Aluminum Substitution and C-A-S-H Formation}
\rnote{This section was rewritten after the first rejection. The first draft built it on figures from a Science Advances page the gate could not reach; the accepted version uses the newly opened Risk Frontiers [7] and Berkeley Lab [4] pages.}
One of the most significant discoveries in Roman concrete research concerns the binder chemistry. In Portland cement concrete, the primary binding compound is calcium-silicate-hydrate (C-S-H). Roman concrete, however, produces a significantly different compound: calcium-aluminum-silicate-hydrate (C-A-S-H) \rcite{1}.

\rtrace{turn 7 \texttt{open(url)} \(\rightarrow\) turn 8 \texttt{open(id=page\_img\_70)}}{Source page: \emph{Why is Roman concrete more durable than modern concrete?}; caption on the page: ``Figure 4: SEM image showing the presence of aluminous tobermorite and phillipsite within Roman marine concrete. Image from Jackson et. al.,\ldots''\par\textbf{Why it was used.} SEM evidence for the two minerals that the section names.}
\rimagebox{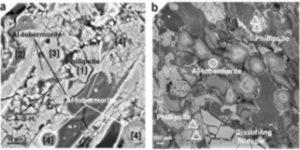}{SEM image showing the presence of aluminous tobermorite and phillipsite within Roman marine concrete}{\rbox{0.19}{0.35}{0.41}{0.47}{}\rbox{0.83}{0.73}{0.99}{0.83}{}}
The scanning electron microscope image reveals the presence of aluminous tobermorite and phillipsite within Roman marine concrete \rcite{7}.\rimgref{5} These minerals form through low-temperature water-rock reactions as seawater percolates through the massive structures, reinforcing the cementing matrix in a kind of regenerative process that strengthens the concrete \rcite{4}.

\rtrace{turn 7 \texttt{open(url)} \(\rightarrow\) turn 8 \texttt{open(id=page\_img\_69)}}{Source page: \emph{Why is Roman concrete more durable than modern concrete?}; caption on the page: ``Figure 3: The research ground lead by Marie Jackson obtaining samples from the Portus Cosanus pier in Orbetello Italy. Image: Marie Jackson.''\par\textbf{Why it was used.} Sampling at the Portus Cosanus pier, supporting the sentence on drilled samples.}
\rimagebox{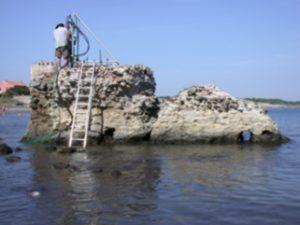}{The research ground led by Marie Jackson obtaining samples from the Portus Cosanus pier in Orbetello Italy}{}
The research team obtained samples by drilling Roman harbour structures and compared them with volcanic rock \rcite{7}.\rimgref{6} The combination of scanning electron microscopy, micro X-ray diffraction, Raman spectroscopy, and electron probe microanalysis revealed the complex mineralogical composition.

\rsec{Al-Tobermorite: The Key to Durability}
Perhaps the most remarkable discovery in Roman concrete research is the presence of Al-tobermorite, a rare mineral that does not occur in conventional modern concrete but appears routinely in the mortars of ancient seawater concretes \rcite{1}.

\rtrace{turn 2 \texttt{open(url)} \(\rightarrow\) turn 3 \texttt{open(id=page\_img\_8)}}{Source page: \emph{Learning from Roman Seawater Concrete}; caption on the page: ``Left: Integrated high-pressure powder x-ray diffraction patterns of Al-tobermorite obtained at Beamline 12.2.2. The Al-tobermorite specimen\ldots''\par\textbf{Why it was used.} The pressure series on the left is the evidence for the stiffness claim in the next sentence.}
\rimagebox{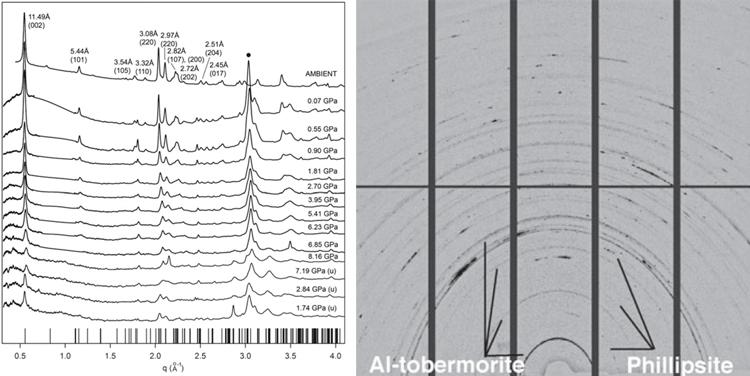}{High-pressure X-ray diffraction patterns of Al-tobermorite and Debye rings from microdiffraction experiments}{\rbox{0.01}{0.08}{0.42}{0.97}{}}
The X-ray diffraction patterns show Al-tobermorite's crystal structure under varying pressures, demonstrating its mechanical stability \rcite{1}.\rimgref{7} High-pressure X-ray diffraction measurements revealed that \fact{Al-tobermorite has greater stiffness than poorly crystalline C-A-S-H}, providing a model for concrete strength and durability \rcite{1}.

\rtrace{turn 7 \texttt{open(url)} \(\rightarrow\) turn 8 \texttt{open(id=page\_img\_45)}}{Source page: \emph{Ancient Roman Concrete Stands the Test of Time - ASME}; caption on the page: ``Image''\par\textbf{Why it was used.} The page's caption is only the placeholder ``Image''; the report's caption repeats the article's subtitle.}
\rimagebox{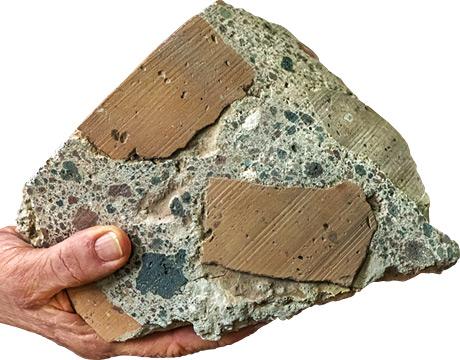}{The ancient Romans had a recipe for concrete that won't corrode in seawater}{}
The ancient Romans had a recipe for concrete that won't corrode in seawater.\rimgref{8} Scientists are trying to figure it out, and the answer lies in the unique mineral chemistry of the Roman formulation \rcite{5}.

\rsec{Self-Healing Mechanisms: The Role of Lime Clasts}
Recent research has fundamentally changed understanding of lime clasts in Roman concrete. Previously dismissed as evidence of sloppy mixing or poor-quality raw materials, these millimeter-scale bright white features are now recognized as purposefully incorporated self-healing agents \rcite{2}.

\rtrace{turn 2 \texttt{open(url)} \(\rightarrow\) turn 3 \texttt{open(id=page\_img\_21)}}{Source page: \emph{Riddle solved: Why was Roman concrete so durable? | MIT News | Massachusetts Institute of Technology}; caption on the page: ``On left, Roman concrete structures of Privernum, Italy with icy puddles, orange sky, and rural location. On right, the concrete fragment is\ldots''\par\textbf{Why it was used.} The image that shows the lime clast: the red region of the color-mapped fragment.}
\rimagebox{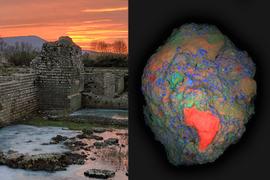}{Roman concrete structures at Privernum with colorized fragment showing lime clast in red}{\rbox{0.67}{0.17}{0.83}{0.42}{}}
The colorized concrete fragment clearly shows the lime clast (red) within the matrix \rcite{2}.\rimgref{9} These macroscopic inclusions serve as critical sources of reactive calcium for long-term pore and crack-filling or post-pozzolanic reactivity within the cementitious constructs \rcite{2}.

\rtrace{turn 2 \texttt{open(url)} \(\rightarrow\) turn 3 \texttt{open(id=page\_img\_22)}}{Source page: \emph{Riddle solved: Why was Roman concrete so durable? | MIT News | Massachusetts Institute of Technology}; caption on the page: ``The concrete fragment is colorized with rainbow colors, including a prominent section colored red.''\par\textbf{Why it was used.} The report says this fragment shows a prominent red lime-clast region. The description follows the caption on the page; in this crop the red region is not visible.}
\rimagebox{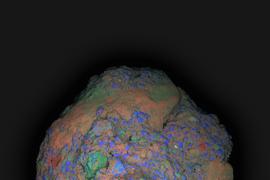}{Colorized concrete fragment with prominent red lime clast region}{}
The detailed colorization reveals how the lime clast integrates with the surrounding matrix \rcite{2}.\rimgref{10} When cracks form, seawater mobilizes calcium from these clasts, triggering reactive hydration that heals the cracks and remodels the interfacial zones \rcite{2}.

\rtrace{turn 2 \texttt{open(url)} \(\rightarrow\) turn 3 \texttt{open(id=page\_img\_25)}}{Source page: \emph{Riddle solved: Why was Roman concrete so durable? | MIT News | Massachusetts Institute of Technology}; caption on the page: ``cement chemistry''\par\textbf{Why it was used.} Image check. The report describes a hot-mixing diagram with clast core, hydration rim and carbonation. The image shows a highway overpass with a color-mapped micrograph inset, and the page's alt text is only ``cement chemistry''. The description is not grounded in the image.}
\rimagebox{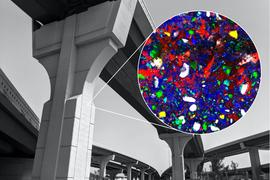}{Cement chemistry diagram showing hot mixing process and lime clast formation}{}
The cement chemistry diagram illustrates the hot mixing process: volcanic ash, sand, aggregate (tephra), water, and quicklime are mixed, with the exothermic reaction creating lime clasts \rcite{2}.\rimgref{11} The hydration and expansion process forms the clast core, hydration rim, and eventually carbonation products.

\rtrace{turn 2 \texttt{open(url)} \(\rightarrow\) turn 3 \texttt{open(id=page\_img\_26)}}{Source page: \emph{Riddle solved: Why was Roman concrete so durable? | MIT News | Massachusetts Institute of Technology}; caption on the page: ``Assistant Professor Admir Masic uses an x-ray fluorescence spectrometer to analyze ancient materials in Italy. Masic is at the forefront of\ldots''\par\textbf{Why it was used.} Shows X-ray fluorescence analysis in the field, supporting the sentence on Masic's method.}
\rimagebox{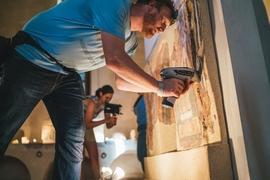}{Admir Masic using X-ray fluorescence spectrometer to analyze ancient materials}{}
MIT Assistant Professor Admir Masic employs X-ray fluorescence spectrometry to analyze ancient materials, pioneering an ``antiqua-inspired'' approach to new materials development \rcite{2}.\rimgref{12} This research demonstrates that modern lime clast-containing cementitious mixtures exhibit self-healing potential, paving the way for more durable, resilient, and sustainable concrete formulations \rcite{2}.

\rsec{Environmental and Sustainability Implications}
\rnote{Reference check. The figure of up to one ton of CO\textsubscript{2} per ton of cement is not on the Kilgore page [6]. The first draft cited it to the Science Advances paper, which the gate could not reach.}
The Roman concrete recipe offers significant environmental advantages over modern Portland cement. The Roman formulation required \fact{approximately 10\% lime by weight, made at two-thirds or less the temperature required by Portland cement} \rcite{1}. This translates to substantially lower CO\textsubscript{2} emissions, as Portland cement production alone releases up to 1 metric ton of CO\textsubscript{2} per metric ton of material produced \rcite{6}.

\rtrace{turn 7 \texttt{open(url)} \(\rightarrow\) turn 8 \texttt{open(id=page\_img\_47)}}{Source page: \emph{Ancient Roman Concrete Stands the Test of Time - ASME}; caption on the page: ``The walls of Trajan's Markets in Rome have stood the test of time and the elements for nearly 2,000 years. Image: Marie Jackson / Berkeley L\ldots''\par\textbf{Why it was used.} Trajan's Markets, used to show that the durability extends beyond marine structures.}
\rimagebox{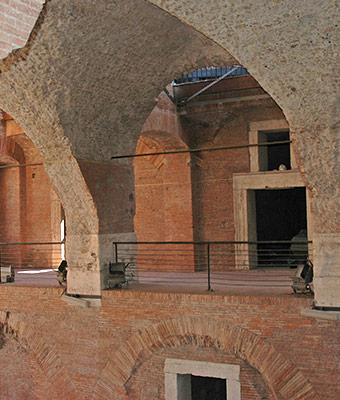}{The walls of Trajan's Markets in Rome have stood the test of time and the elements for nearly 2,000 years}{}
The walls of Trajan's Markets in Rome have stood the test of time and the elements for nearly 2,000 years \rcite{5}.\rimgref{13} This demonstrates the long-term durability of Roman concrete technology in various environmental conditions.

\rsec{Comparison with Modern Concrete}
Modern concrete consists of Portland cement, stone aggregates, and sand mixed with fresh water \rcite{6}. The main advantage of modern concrete is that it sets and dries quickly compared to Roman concrete. Modern concrete is also widely available and accessible in mass production \rcite{6}.

However, modern concrete has significant disadvantages. Modern concrete is likely to crack because of the reinforcement steel used in the mixture. \fact{The steel corrosion inside concrete can cause the concrete to expand and crack} \rcite{6}. Another disadvantage is that modern Portland cement production emits about 8\% of carbon dioxide into the atmosphere \rcite{6}.

Roman concrete is more environmentally friendly because production does not emit toxic gases \rcite{6}. This concrete is less likely to crack under stress due to less water in the concrete mixture with no reinforcement steels \rcite{6}.

\rsec{Conclusion}
Roman maritime concrete demonstrates superior long-term durability compared to modern Portland cement formulations due to several interconnected factors: (1) the use of volcanic ash with silica and alumina content that enables pozzolanic reactivity; (2) the formation of C-A-S-H binder with aluminum substitution for silicon; (3) the crystallization of Al-tobermorite, which provides exceptional stiffness and stability; and (4) the purposeful inclusion of lime clasts that enable self-healing through reactive calcium mobilization. These ancient materials systems, developed through empirical optimization over centuries, offer sustainable prototypes for modern concrete engineering. By incorporating lime clasts and volcanic ash into modern formulations, engineers can develop more durable, resilient, and environmentally friendly concrete that reduces the carbon footprint of construction while extending infrastructure service life.

\rsec{References}
\rnote{Reference [8] is not cited in the text. It was added after the second rejection, which asked for an eighth accessible source.}
\rcite{1} Learning from Roman Seawater Concrete\rurl{https://als.lbl.gov/learning-from-roman-concrete/} - ALS Berkeley

\rcite{2} Riddle solved: Why was Roman concrete so durable?\rurl{https://news.mit.edu/2023/roman-concrete-durability-lime-casts-0106} - MIT News

\rcite{3} Roman Concrete\rurl{https://ancientengrtech.wisc.edu/roman-concrete/} - UW-Madison Ancient Engineering Technologies

\rcite{4} The Ancient Roman Secret to Concrete Resilience in Seawater\rurl{https://als.lbl.gov/ancient-roman-secret-concrete-resilience-seawater/} - ALS Berkeley

\rcite{5} Ancient Roman Concrete Stands the Test of Time\rurl{https://www.asme.org/topics-resources/content/ancient-roman-concrete-stands-test-time} - ASME

\rcite{6} Roman Concrete vs. Modern Concrete\rurl{https://kilgorecompanies.com/blog/2022/08/26/roman-concrete-vs-modern-concrete/} - Kilgore Companies

\rcite{7} Why is Roman concrete more durable than modern concrete?\rurl{https://riskfrontiers.com/insights/why-is-roman-concrete-more-durable-than-modern-concrete/} - Risk Frontiers

\rcite{8} Ancient Roman concrete found to have self-healing capability\rurl{https://www.iom3.org/resource/roman-concrete.html} - IOM3

\rimgsources{\rimgsrc{Images~1, 7}{Learning from Roman Seawater Concrete}{https://als.lbl.gov/learning-from-roman-concrete/}
\rimgsrc{Images~2, 8, 13}{Ancient Roman Concrete Stands the Test of Time - ASME}{https://www.asme.org/topics-resources/content/ancient-roman-concrete-stands-test-time}
\rimgsrc{Images~3\textendash{}6}{Why is Roman concrete more durable than modern concrete?}{https://riskfrontiers.com/insights/why-is-roman-concrete-more-durable-than-modern-concrete/}
\rimgsrc{Images~9\textendash{}12}{Riddle solved: Why was Roman concrete so durable? | MIT News | Massachusetts Institute of Technology}{https://news.mit.edu/2023/roman-concrete-durability-lime-casts-0106}}
\rgate{\item Submission 1 (turn 5): {\color{rframe}\textbf{WARN}} --- 8,516 chars, 17~images, 7~sources; \texttt{source\_count}, \texttt{reference\_unreachable}\,\(\times\)4, \texttt{accessible\_reference\_count}
\item Submission 2 (turn 9): {\color{rframe}\textbf{WARN}} --- 8,069 chars, 13~images, 7~sources; \texttt{source\_count}, \texttt{accessible\_reference\_count}
\item Submission 3 (turn 12): {\color{tok}\textbf{OK}} --- 8,063 chars, 13~images, 8~sources}

\end{reportcard}

\FloatBarrier
\subsection{Concise and Detail Reports for the Desert-Plant Photosynthesis Request}
\label{app:examples-cam}

The second request asks how crassulacean acid metabolism (CAM), a form of
photosynthesis that takes up carbon dioxide at night, evolved in unrelated
desert plants. It shows the same difference between the two modes on an
unrelated topic. The Concise-mode report answers in 1,765 characters with
2 images and 5 sources, under three sections on anatomy, physiology, and
metabolism. The Detail-mode report uses 7,377 characters, 13 images, and
9 sources, and adds separate sections on carbon isotope ratios, nocturnal gas
exchange, facultative CAM, and phylogenetic distribution. Both reports pass
the gate on the first submission and are annotated as in
Appendix~\ref{app:examples-reports}. Most sources are PubMed Central pages,
which now refuse automated access, so their claims were checked against the
page text and figure captions captured during the run and against the image
pixels, not against a fresh copy of each page.

\begin{reportcard}{CONCISE}{1,765 chars \textperiodcentered{} 2 images \textperiodcentered{} 5 sources \textperiodcentered{} 1 submission}{How has CAM photosynthesis diversified across unrelated desert plant lineages, and what do anatomical cross-section micrographs, carbon isotope ratio charts, and nocturnal gas exchange diagrams from public herbarium and research collections reveal about the convergent evolutionary solutions to arid environments in agaves, cacti, and euphorbias?}
\rtitle{Convergent CAM Photosynthesis in Desert Plants}
\clue{Crassulacean acid metabolism (CAM)} has evolved independently across diverse plant lineages as a convergent solution to \clue{arid environments}. The Agavoideae subfamily exemplifies this diversity, with \fact{species ranging from C3 to CAM and intermediate phenotypes} \rcite{1}.

\rsec{Anatomical Convergence}
\rnote{Reference check. The figure caption on page [1] labels the undifferentiated mesophyll (M), intercellular airspace (IAS) and vascular bundles (VB) of \emph{Yucca aloifolia}; ``large vacuoles for malate storage'' and ``reduced'' airspace are not in that caption. Page [2] says that unrelated groups evolved desert traits independently and that euphorbias and cacti are stem succulents, but it does not use the phrase ``water-storing tissues''.}
Leaf \clue{cross-sections} reveal shared structural adaptations. CAM species like \emph{Yucca aloifolia} show \fact{undifferentiated mesophyll (M)} with large vacuoles for malate storage, reduced \fact{intercellular air spaces (IAS)}, and prominent \fact{vascular bundles (VB)} \rcite{1}.\rimgref{1} These features appear across unrelated succulents including \clue{agaves, cacti, and euphorbias}, which \fact{independently evolved} fleshy, water-storing tissues \rcite{2}.

\rtrace{turn 3 \texttt{open(url)} \(\rightarrow\) turn 4 \texttt{open(id=page\_img\_23)}}{Source page: \emph{The Agavoideae: an emergent model clade for CAM evolutionary biology}; caption on the page: ``(A) Anatomical cross-sections of leaves from Yucca aloifolia (top, CAM), Y. gloriosa (middle, C3 + CAM) and Y. filamentosa (bottom, C3). VB: vascular bundle, M: undifferentiated mesophyll, IAS: intercellular airspace''\par\textbf{Why it was used.} The only micrograph the run opened, and the one that puts a CAM, an intermediate and a C3 leaf side by side. The text relies on the top-left section (CAM).}
\rimagebox{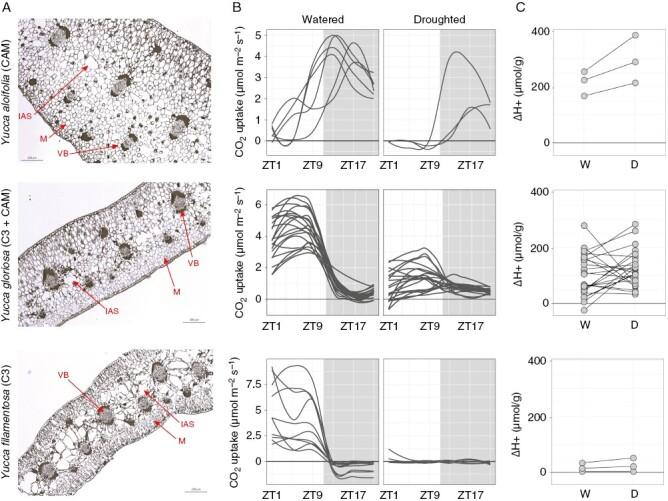}{Anatomical cross-sections of Yucca leaves showing CAM, C3+CAM, and C3 phenotypes with labeled vascular bundles, mesophyll, and intercellular air spaces}{\rbox{0.0}{0.67}{0.31}{1.0}{}}
\rsec{Physiological and Biochemical Evidence}
\rnote{Reference check. The last sentence has the direction reversed. Stronger CAM moves \(\delta^{13}\)C toward \emph{less} negative values, because the nocturnal carboxylase discriminates less against \textsuperscript{13}C; values near \(-\)25 to \(-\)30\textperthousand{} are typical of C3 plants. The page text captured from [4] gives no ranges, and its \(\delta^{13}\)C chart, opened in turn 4 but not used, does not label photosynthetic types. The model's reasoning before submission already contains the error; the gate does not check claims.}
\clue{Nocturnal gas exchange} diagrams demonstrate the functional convergence: \fact{CAM plants open stomata at night to fix CO\textsubscript{2}, storing it as malic acid} for daytime photosynthesis \rcite{3}.\rimgref{2} This inverted rhythm minimizes water loss during hot days. \clue{Carbon isotope ratios} (\(\delta\)\textsuperscript{1}\textsuperscript{3}C) further confirm CAM evolution across lineages, with values shifting toward more negative values (\(\sim\)--25 to --30\textperthousand{}) as plants adopt stronger CAM phenotypes \rcite{4}.

\rtrace{turn 3 \texttt{open(url)} \(\rightarrow\) turn 4 \texttt{open(id=page\_img\_106)}}{Source page: \emph{Stomatal Biology of CAM Plants}; caption on the page: ``Summary of differences in stomatal biology of C3 and CAM plants. A, Representative curves of day/night rhythms in stomatal conductance (gs) in C3 and CAM plants.''\par\textbf{Why it was used.} The run's only diagram of stomatal opening over a day and night; the CAM curve (top right) is high at night and closed at midday. The text never names this page, so the renderer listed it under the image sources.}
\rimagebox{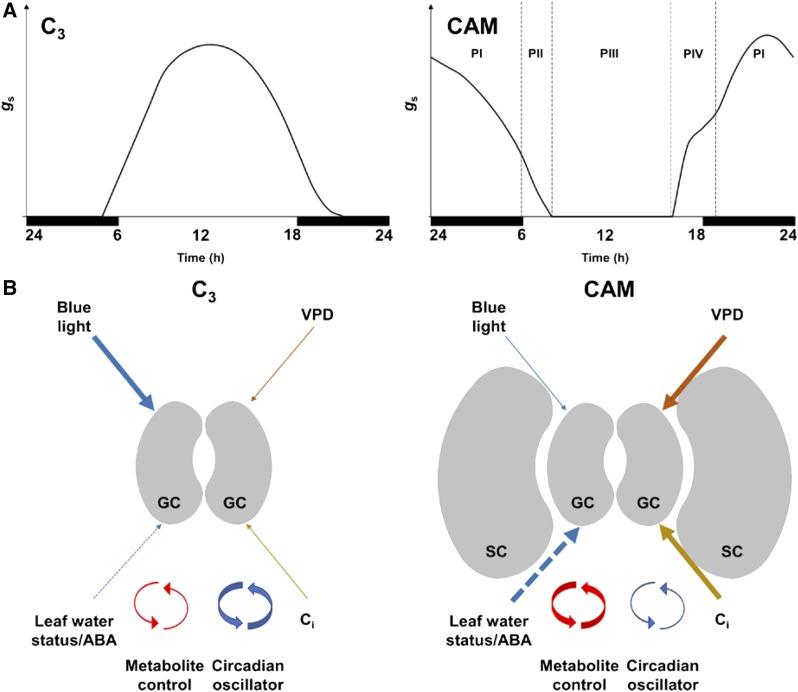}{Stomatal conductance rhythms comparing C3 and CAM plants, showing nocturnal opening in CAM species}{\rbox{0.5}{0.73}{1.0}{1.0}{}}
\rsec{Metabolic Integration}
The CAM cycle requires elevated nocturnal respiration to power malate accumulation. \fact{Flux balance models and physiological data} show mitochondria must \fact{increase respiratory oxygen consumption at night} to generate ATP and NADH for phosphoenolpyruvate carboxylase activity \rcite{3}. This metabolic integration---combining nocturnal CO\textsubscript{2} uptake, vacuolar storage, and daytime decarboxylation---represents a convergent biochemical solution to water limitation that has arisen repeatedly across \fact{angiosperms, gymnosperms, and ferns} \rcite{4}.

\rsec{References}
\begin{rlist}
\item \rcite{1} The Agavoideae: an emergent model clade for CAM evolutionary biology\rurl{https://pmc.ncbi.nlm.nih.gov/articles/PMC10799990/}
\item \rcite{2} Convergent Evolution of Desert Plants from Africa and the Americas\rurl{https://www.csustan.edu/biology/stan-state-greenhouse/convergent-evolution-desert-plants-africa-and-americas}
\item \rcite{3} Elevated nocturnal respiratory rates in the mitochondria of CAM plants\rurl{https://pmc.ncbi.nlm.nih.gov/articles/PMC10799998/}
\item \rcite{4} The diverse diaspora of CAM: a pole-to-pole sketch\rurl{https://pmc.ncbi.nlm.nih.gov/articles/PMC10800000/}
\end{rlist}
\rimgsources{\rimgsrc{Image~1}{The Agavoideae: an emergent model clade for CAM evolutionary biology}{https://pmc.ncbi.nlm.nih.gov/articles/PMC10799990/}
\rimgsrc{Image~2}{Stomatal Biology of CAM Plants}{https://pmc.ncbi.nlm.nih.gov/articles/PMC5462028/}}
\rgate{\item Submission 1 (turn 5): {\color{tok}\textbf{OK}} --- 1,765 chars, 2 images, 5 sources}

\end{reportcard}

\begin{reportcard}{DETAIL}{7,377 chars \textperiodcentered{} 13 images \textperiodcentered{} 9 sources \textperiodcentered{} 1 submission}{How has CAM photosynthesis diversified across unrelated desert plant lineages, and what do anatomical cross-section micrographs, carbon isotope ratio charts, and nocturnal gas exchange diagrams from public herbarium and research collections reveal about the convergent evolutionary solutions to arid environments in agaves, cacti, and euphorbias?}
\rtitle{Convergent Evolution of CAM Photosynthesis in Desert Plants: Evidence from Anatomy, Isotopes, and Gas Exchange}
\rsec{Introduction}
\clue{Crassulacean acid metabolism (CAM)} photosynthesis has \fact{evolved independently over 60 times across vascular plants, making it one of the most striking examples of evolutionary convergence on a complex trait} \rcite{1}. This pathway allows plants to open stomata at night when transpirational water loss is minimized, fixing CO\textsubscript{2} into malic acid for daytime use. The Agavoideae (\clue{agaves} and yuccas), Cactaceae (\clue{cacti}), and Euphorbiaceae (\clue{euphorbias}) represent three unrelated lineages that have independently evolved CAM and succulence as solutions to \clue{arid environments}. This report examines how \clue{anatomical cross-sections}, \clue{carbon isotope} \clue{ratios}, and \clue{nocturnal gas exchange} data reveal convergent evolutionary solutions across these lineages.

\rsec{CAM as a Convergent Adaptation to Aridity}
\rnote{Reference check. The epiphytes in the paragraph after Image~1 come from the figure caption on page [3] (an orchid, two bromeliads including Spanish moss); ``leafless photosynthetic roots'' is not in the page text or caption captured during the run.}
CAM photosynthesis is \fact{present in approximately 5\% of vascular plants}, distributed across \fact{ferns, gymnosperms, and angiosperms} \rcite{2}. The pathway involves \fact{nocturnal CO\textsubscript{2} fixation via phosphoenolpyruvate carboxylase (PEPC), storage of malic acid in vacuoles, and daytime decarboxylation} to supply CO\textsubscript{2} to Rubisco behind closed stomata \rcite{3}. This temporal separation of carbon acquisition and fixation dramatically improves water-use efficiency, enabling survival in deserts, epiphytic habitats, and other water-limited niches.

\rtrace{turn 8 \texttt{open(url)} \(\rightarrow\) turn 9 \texttt{open(id=page\_img\_215)}}{Source page: \emph{Prospects and perspectives: inferring physiological and regulatory targets for CAM from molecular and modelling approaches}; caption on the page: ``Diversity in succulent CAM lifeforms and a typical gas exchange profile. Terrestrial CAM plants: (A) massive stem succulent convergence in cacti (left) and euphorbia (right) species, with agave (centre)''\par\textbf{Why it was used.} The one photograph in the run with a cactus, a euphorbia and an agave side by side (panel A); panel I adds the four CAM phases over a day.}
\rimagebox{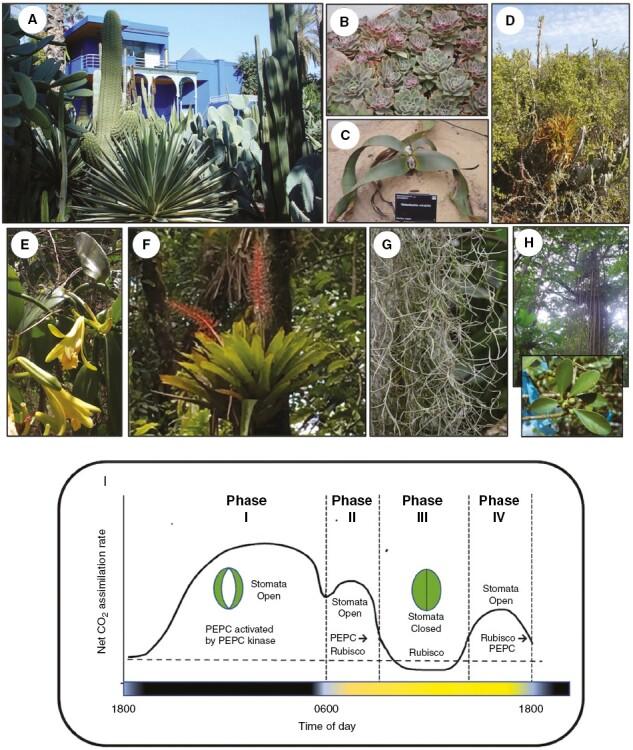}{Diversity of succulent CAM lifeforms including cacti, euphorbias, agaves, and epiphytic species, with a schematic of CAM phases showing nocturnal stomatal opening and daytime CO\textsubscript{2} release}{\rbox{0.0}{0.71}{0.51}{1.0}{}}
The morphological diversity of CAM plants is remarkable. Terrestrial representatives include \fact{massive stem succulents like cacti and euphorbias, leaf succulents like agaves, and more herbaceous forms like Crassulaceae} \rcite{3}.\rimgref{1} Epiphytic CAM plants include orchids, bromeliads, and Spanish moss. This diversity demonstrates that CAM can be integrated into many body plans, from arborescent forms to leafless photosynthetic roots.

\rsec{Anatomical Convergence: Leaf Structure and Succulence}
\rnote{Reference check. The figure on page [4] labels the vascular bundles, mesophyll and intercellular airspace of the three \emph{Yucca} leaves; it does not say that the CAM leaves are thicker or that their airspaces are large.}
Anatomical cross-sections reveal convergent structural adaptations for CAM function. The Agavoideae subfamily provides an excellent model for studying CAM evolution because it \fact{contains species spanning the C3--CAM continuum} \rcite{4}.

\rtrace{turn 2 \texttt{open(url)} \(\rightarrow\) turn 3 \texttt{open(id=page\_img\_23)}}{Source page: \emph{The Agavoideae: an emergent model clade for CAM evolutionary biology}; caption on the page: ``(A) Anatomical cross-sections of leaves from Yucca aloifolia (top, CAM), Y. gloriosa (middle, C3 + CAM) and Y. filamentosa (bottom, C3). VB: vascular bundle, M: undifferentiated mesophyll, IAS: intercellular airspace''\par\textbf{Why it was used.} The same figure as in the Concise report: a CAM, an intermediate and a C3 leaf from one genus. The text relies on the three sections in the left column.}
\rimagebox{figures/showcase/cam/page_img_23__mcad062_fig3__9018d4c929.jpg}{Anatomical cross-sections of Yucca leaves showing CAM (Y. aloifolia), C3+CAM intermediate (Y. gloriosa), and C3 (Y. filamentosa) phenotypes with labeled vascular bundles, mesophyll, and intercellular air spaces}{\rbox{0.0}{0.0}{0.31}{1.0}{}}
In \emph{Yucca}, CAM species like \emph{Y. aloifolia} exhibit thick leaves with large intercellular air spaces and \fact{undifferentiated mesophyll}, while C3 species like \emph{Y. filamentosa} show more typical leaf anatomy \rcite{4}.\rimgref{2} The hybrid \emph{Y. gloriosa} \fact{displays intermediate anatomy and physiology, making it a valuable system for studying CAM evolution} \rcite{8}.\rimgref{3}

\rtrace{turn 8 \texttt{open(url)} \(\rightarrow\) turn 9 \texttt{open(id=page\_img\_240)}}{Source page: \emph{Gas exchange and leaf anatomy of a C3--CAM hybrid, Yucca gloriosa (Asparagaceae)}; caption on the page: ``Cross sections of Y. aloifolia (A and B), Y. gloriosa (C and D), and Y. filamentosa (E and F) at 5\(\times\) (A, C, E) and 10\(\times\) (B, D, F) magnification.''\par\textbf{Why it was used.} Shows the hybrid's leaf between its two parents at two magnifications, for the sentence on \emph{Y. gloriosa}.}
\rimagebox{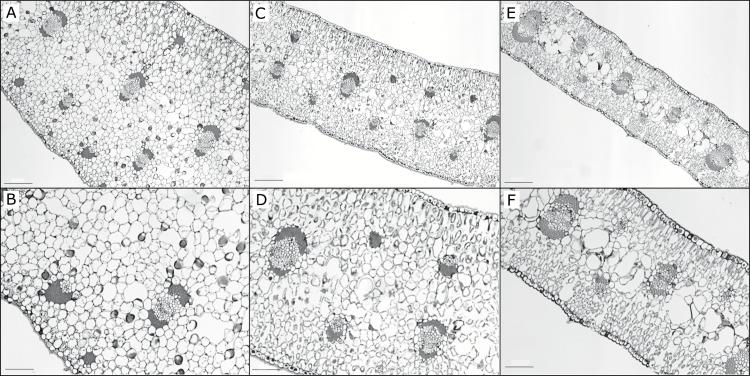}{Cross-sections of Y. aloifolia (CAM), Y. gloriosa (C3+CAM), and Y. filamentosa (C3) at 5\(\times\) and 10\(\times\) magnification showing progressive anatomical differences}{}
The \emph{Euphorbia milii} leaf cross-section reveals another convergent anatomical solution. The leaf contains \fact{palisade parenchyma, spongy parenchyma, and a bundle sheath with chloroplasts, suggesting possible C\textsubscript{2} photosynthesis} alongside \fact{weak CAM} \rcite{5}.\rimgref{4}

\rtrace{turn 5 \texttt{open(url)} \(\rightarrow\) turn 6 \texttt{open(id=page\_img\_139)}}{Source page: \emph{Crassulacean acid metabolism-cycling in Euphorbia milii}; caption on the page: ``Cross-sections of the leaf of E. milii. UE, upper epidermis; PP, palisade parenchyma; VB, vascular bundle; BS, bundle sheath; SP, spongy parenchyma; LE, lower epidermis. Arrowheads point at chloroplasts.''\par\textbf{Why it was used.} The run's only cross-section from a euphorbia, one of the three lineages in the request.}
\rimagebox{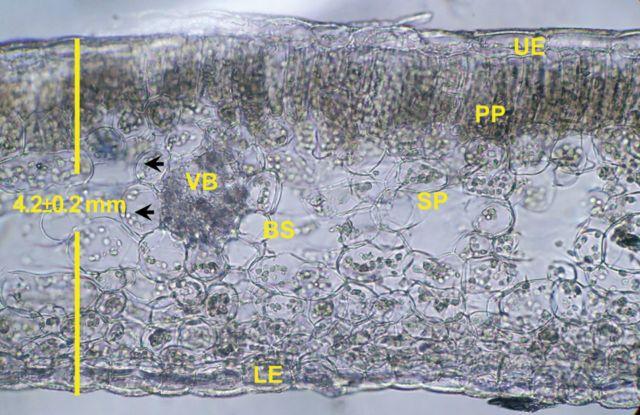}{Cross-section of E. milii leaf showing upper epidermis, palisade parenchyma, vascular bundle, bundle sheath, spongy parenchyma, and lower epidermis}{}
\fact{Succulence}---fleshy tissue with high water content---\fact{is hypothesized to be essential for CAM function because it allows storage of large amounts of assimilated carbon as malate and assists in trapping CO\textsubscript{2} released during the day} \rcite{1}.

\rsec{Carbon Isotope Ratios: Quantifying CAM Contribution}
\rnote{Reference check. Page [6] states that PEP carboxylase discriminates less against \textsuperscript{13}C and that strong CAM plants resemble C4 plants, but the ranges in parentheses are not in the text captured from it. The peaks at \(-\)13\textperthousand{} and \(-\)26\textperthousand{} that the report gives for the 506-species figure come from a bromeliad survey described elsewhere on the page; the figure itself peaks near \(-\)14\textperthousand{} and \(-\)28\textperthousand{}. Page [2] does not say which Aizoaceae subfamilies are CAM.}
Carbon isotope ratios (\(\delta\)\textsuperscript{1}\textsuperscript{3}C) provide a powerful tool for identifying CAM plants and quantifying the relative contribution of nocturnal versus diurnal carbon fixation. The enzyme \fact{PEP carboxylase discriminates less against \textsuperscript{1}\textsuperscript{3}C than Rubisco, so plants with strong CAM exhibit \(\delta\)\textsuperscript{1}\textsuperscript{3}C values similar to C4 plants} (around --10\textperthousand{} to --15\textperthousand{}), while C3 plants show more negative values (around --25\textperthousand{} to --30\textperthousand{}) \rcite{6}.

\rtrace{turn 2 \texttt{open(url)} \(\rightarrow\) turn 3 \texttt{open(id=page\_img\_78)}}{Source page: \emph{How Closely Do the \(\delta^{13}\)C Values of Crassulacean Acid Metabolism Plants Reflect the Proportion of CO\textsubscript{2} Fixed during Day and Night?}; caption on the page: ``Relationships between the cumulative net carbon gain in the light and the dark and the \(\delta^{13}\)C values \ldots{} Extrapolation of the fitted regression \ldots{} in B indicates \(\delta^{13}\)C values of \(-\)8.7\textperthousand{} for 100\% dark CO\textsubscript{2} uptake and \(-\)26.9\textperthousand{} for 100\% CO\textsubscript{2} uptake during the light.''\par\textbf{Why it was used.} The source of the two end-point values in the next paragraph, which come from panel B.}
\rimagebox{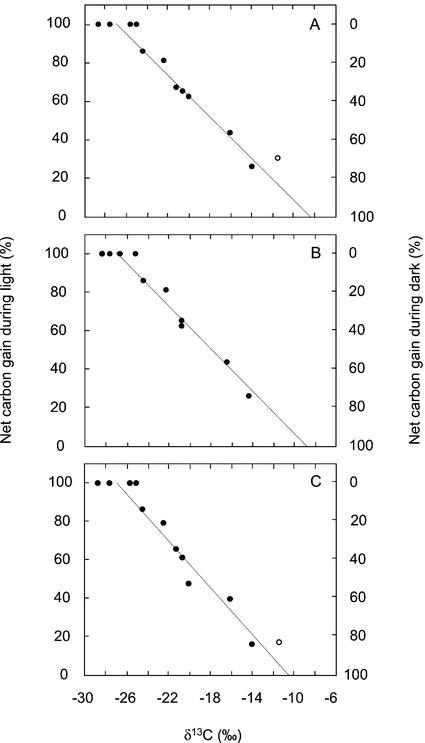}{Relationships between cumulative net carbon gain in light and dark and \(\delta\)\textsuperscript{1}\textsuperscript{3}C values, showing linear correlations for CAM species}{\rbox{0.06}{0.40}{0.95}{0.70}{}}
Experimental studies demonstrate that \fact{\(\delta\)\textsuperscript{1}\textsuperscript{3}C values correlate linearly with the proportion of carbon gained in the dark versus light}. For CAM species, extrapolation indicates \(\delta\)\textsuperscript{1}\textsuperscript{3}C values of approximately \fact{--8.7\textperthousand{} for 100\% dark CO\textsubscript{2} uptake and --26.9\textperthousand{} for 100\% light CO\textsubscript{2} uptake} \rcite{6}.\rimgref{5} This relationship allows researchers to estimate the CAM contribution in natural populations.

\rtrace{turn 2 \texttt{open(url)} \(\rightarrow\) turn 3 \texttt{open(id=page\_img\_79)}}{Source page: \emph{How Closely Do the \(\delta^{13}\)C Values of Crassulacean Acid Metabolism Plants Reflect the Proportion of CO\textsubscript{2} Fixed during Day and Night?}; caption on the page: ``\(\delta^{13}\)C values of 506 species from the Asclepiadaceae, Bromeliaceae, Crassulaceae, Cucurbitaceae, Didiereaceae, Euphorbiaceae, Orchidaceae, Polypodiaceae, and Vittariaceae plant families known to contain both C3 and CAM members.''\par\textbf{Why it was used.} A two-peaked histogram across nine families, including the Euphorbiaceae, for the sentence on the C3--CAM split.}
\rimagebox{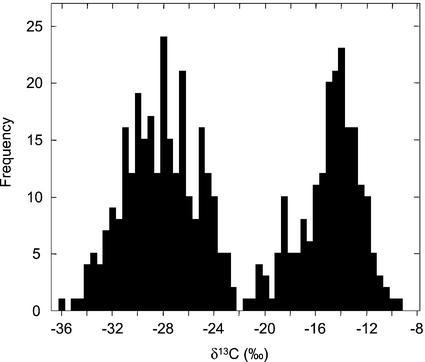}{\(\delta\)\textsuperscript{1}\textsuperscript{3}C values of 506 species from multiple plant families showing bimodal distribution with peaks around --13\textperthousand{} (strong CAM) and --26\textperthousand{} (C3)}{}
The bimodal distribution of \(\delta\)\textsuperscript{1}\textsuperscript{3}C values across plant families reveals the evolutionary transition between C3 and CAM photosynthesis.\rimgref{6} In the Aizoaceae, for example, different subfamilies show distinct \(\delta\)\textsuperscript{1}\textsuperscript{3}C distributions reflecting their CAM status \rcite{2}.\rimgref{7}

\rtrace{turn 5 \texttt{open(url)} \(\rightarrow\) turn 6 \texttt{open(id=page\_img\_116)}}{Source page: \emph{The diverse diaspora of CAM: a pole-to-pole sketch}; caption on the page: ``Values of \(\delta^{13}\)C (per mille) of taxa from the five subfamilies of the Aizoaceae. Note that frequency values include values from different species and multiple measurements of the same species.''\par\textbf{Why it was used.} One family broken down by subfamily, for the Aizoaceae sentence.}
\rimagebox{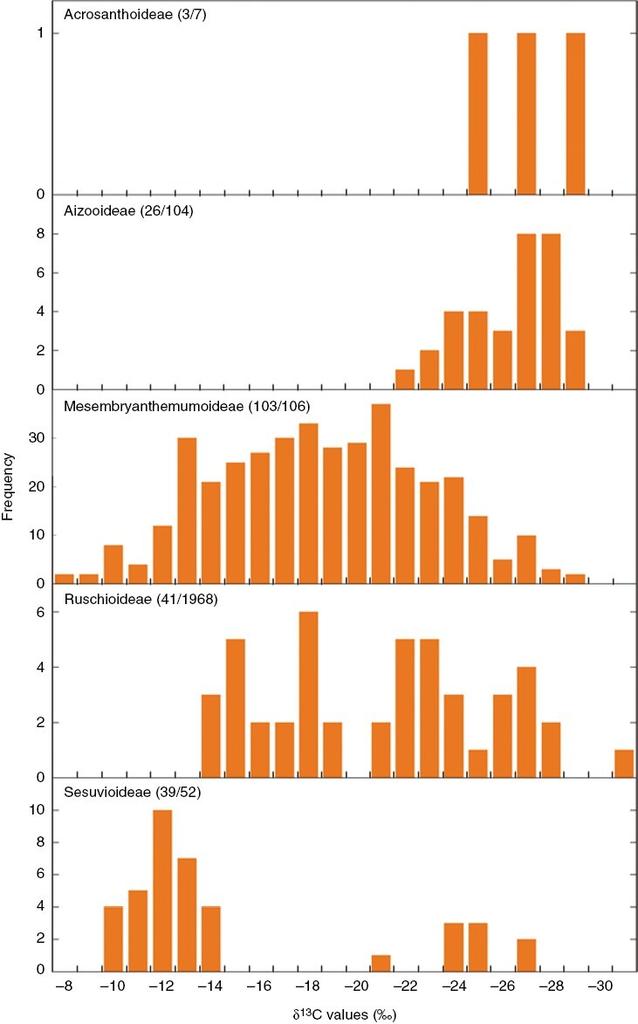}{\(\delta\)\textsuperscript{1}\textsuperscript{3}C values of taxa from five Aizoaceae subfamilies showing variation from C3 to strong CAM}{}
In \emph{Euphorbia milii}, \fact{\(\delta\)\textsuperscript{1}\textsuperscript{3}C values of --25.2\textperthousand{} in leaves and --24.7\textperthousand{} in stems} indicate \fact{weak CAM or CAM-cycling, intermediate between C3 and constitutive CAM plants} \rcite{5}. This intermediate value \fact{results from recycling of nocturnal respiratory CO\textsubscript{2}} rather than direct atmospheric fixation.

\rsec{Nocturnal Gas Exchange: The CAM Rhythm}
Continuous gas exchange measurements reveal the characteristic CAM rhythm of nocturnal CO\textsubscript{2} uptake and daytime release. In \emph{Agave angustifolia}, \fact{CO\textsubscript{2} uptake in the dark contributes about 75\% of daily carbon gain, and this pattern is highly conserved across environmental conditions} \rcite{7}.\rimgref{8}

\rtrace{turn 2 \texttt{open(url)} \(\rightarrow\) turn 3 \texttt{open(id=page\_img\_45)}}{Source page: \emph{Nocturnal versus diurnal CO\textsubscript{2} uptake: how flexible is Agave angustifolia?}; caption on the page: ``Net CO\textsubscript{2} balance of the shoot of a young individual of A. angustifolia during 281 consecutive 12h light/12h dark cycles. Carbon balances are shown for the light and dark periods.''\par\textbf{Why it was used.} The run's only long gas-exchange record from an agave: nine months of day and night carbon balance under changing light and watering.}
\rimagebox{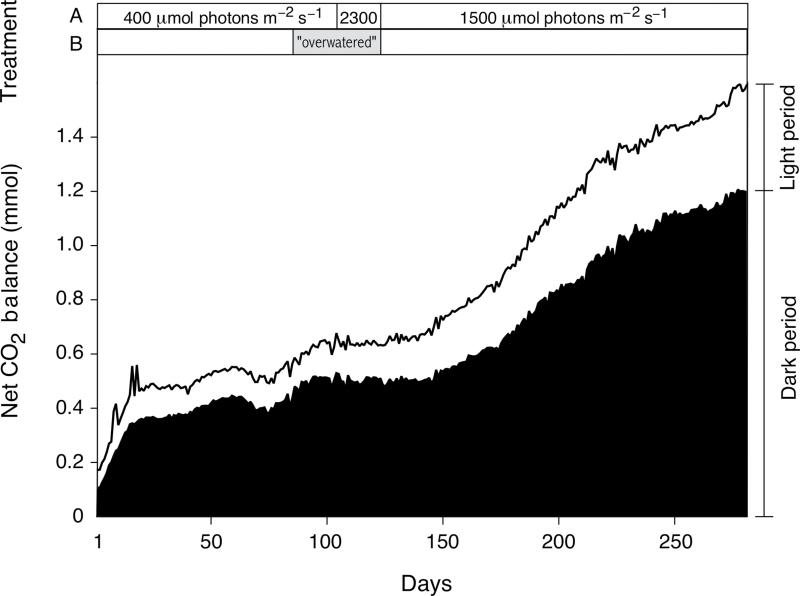}{Net CO\textsubscript{2} balance of young A. angustifolia during 281 consecutive 12h light/12h dark cycles showing consistent nocturnal CO\textsubscript{2} uptake}{}
The nocturnal CO\textsubscript{2} uptake in CAM plants is not simply a matter of stomatal opening. Research on \emph{Clusia} species demonstrates that \fact{CAM requires elevated nocturnal respiratory rates to power vacuolar malic acid accumulation} \rcite{9}.

\rtrace{turn 8 \texttt{open(url)} \(\rightarrow\) turn 9 \texttt{open(id=page\_img\_264)}}{Source page: \emph{Elevated nocturnal respiratory rates in the mitochondria of CAM plants: current knowledge and unanswered questions}; caption on the page: ``Schematic of major metabolite fluxes involved during the nocturnal phase of CAM and their interaction with the respiratory TCA cycle.''\par\textbf{Why it was used.} Shows where respiration feeds malate synthesis at night, for the paragraph on the metabolic cost of CAM.}
\rimagebox{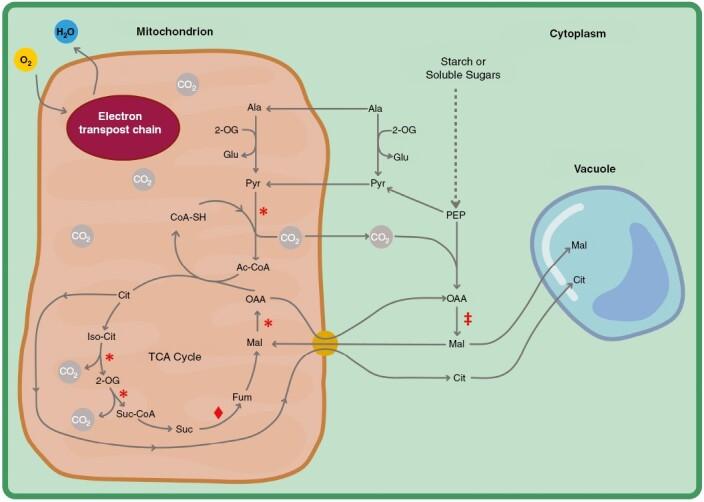}{Schematic of metabolite fluxes during nocturnal CAM phase showing interaction between mitochondrial respiration, TCA cycle, and malate synthesis}{}
The metabolic demands of CAM are substantial. Nocturnal respiration provides the ATP and reducing power needed for PEP regeneration and malate synthesis.\rimgref{9} \fact{Flux balance analysis models predict that CAM requires elevated nocturnal respiratory rates compared to C3 species} \rcite{9}.

\rtrace{turn 8 \texttt{open(url)} \(\rightarrow\) turn 9 \texttt{open(id=page\_img\_266)}}{Source page: \emph{Elevated nocturnal respiratory rates in the mitochondria of CAM plants: current knowledge and unanswered questions}; caption on the page: ``Clusia pratensis (grey dots) and C. tocuchensis (white dots) exhibit facultative CAM and obligate C3 phenotypes, respectively.''\par\textbf{Why it was used.} Pairs a facultative CAM species with a C3 relative under watering and drought, for the sentence on drought-induced CAM.}
\rimagebox{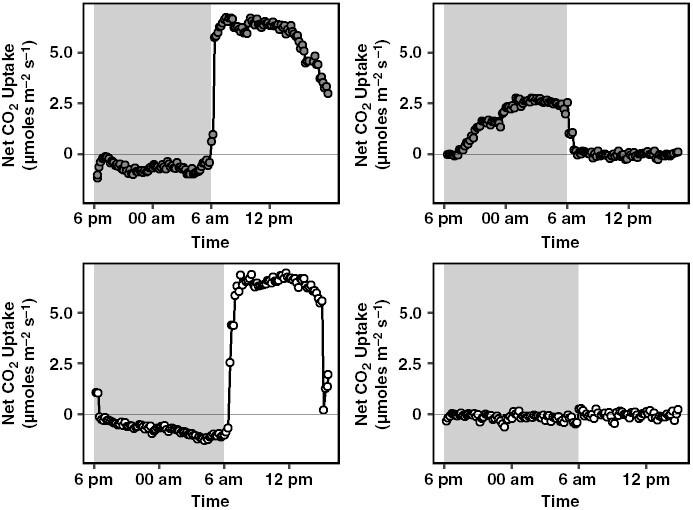}{Diel photosynthetic assimilation rates in well-watered and drought-treated Clusia pratensis (facultative CAM) and C. tocuchensis (obligate C3)}{}
In \fact{facultative CAM} species like \emph{Clusia pratensis}, drought induces CAM, with nocturnal CO\textsubscript{2} uptake becoming prominent under water stress \rcite{9}.\rimgref{10} This plasticity allows plants to optimize carbon gain across varying water availability.

\rsec{Facultative CAM and Environmental Plasticity}
\rnote{Figure check. In the \emph{Lecanopteris} record, drying lowers daytime uptake and adds a small night-time uptake; night uptake never becomes the larger part. In the \emph{E. milii} time course, total acid content rises under drought, but nocturnal accumulation (panel C) does not: at pH 7.0 it stays level, and at pH 8.4 it peaks at day 5 and falls to zero by day 16.}
Not all CAM plants are obligate. Some species, like the ant-fern \emph{Lecanopteris sinuosa}, exhibit \fact{facultative CAM} that can be induced by drought \rcite{2}.

\rtrace{turn 5 \texttt{open(url)} \(\rightarrow\) turn 6 \texttt{open(id=page\_img\_112)}}{Source page: \emph{The diverse diaspora of CAM: a pole-to-pole sketch}; caption on the page: ``Demonstration of facultative CAM in the ant-fern Lecanopteris sinuosa (Polypodiaceae) during 20 days of net CO\textsubscript{2} exchange \ldots{} Net CO\textsubscript{2} exchange during the light is shown as yellow, whereas net CO\textsubscript{2} exchange during the dark is shown as blue.''\par\textbf{Why it was used.} A watering--drying--rewatering cycle in a fern, showing that CAM can switch on and off within weeks.}
\rimagebox{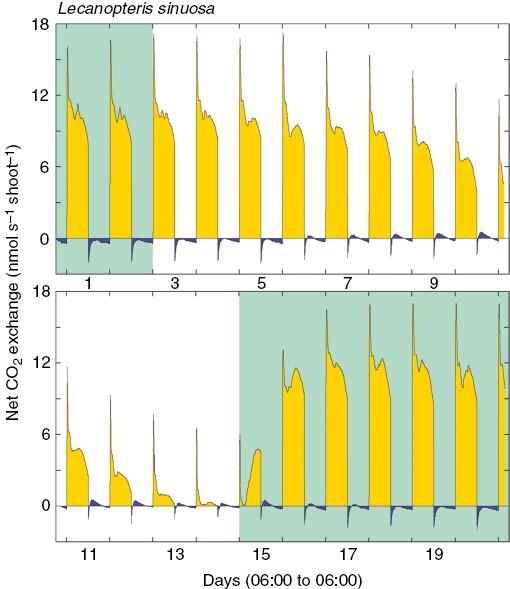}{Demonstration of facultative CAM in Lecanopteris sinuosa during watering-drying-rewatering cycles, with net CO\textsubscript{2} exchange showing nocturnal uptake during drought}{}
In \emph{Lecanopteris sinuosa}, net CO\textsubscript{2} exchange shifts from predominantly diurnal (yellow) to nocturnal (blue) uptake during drought periods, demonstrating the plasticity of CAM expression \rcite{2}.\rimgref{11}

Similarly, \emph{Euphorbia milii} shows \fact{CAM-cycling} rather than strong CAM, with \fact{nocturnal H\textsuperscript{+} accumulation attributable to both malate and citrate} \rcite{5}.

\rtrace{turn 5 \texttt{open(url)} \(\rightarrow\) turn 6 \texttt{open(id=page\_img\_140)}}{Source page: \emph{Crassulacean acid metabolism-cycling in Euphorbia milii}; caption on the page: ``Time course of changes with drought in leaves of E. milii in (A) H\textsuperscript{+} content titrated to pH 7.0 \ldots{} (C) nocturnal H\textsuperscript{+} accumulation \ldots{} and (D) dawn leaf FM per area (circles) and chlorophyll content (triangles).''\par\textbf{Why it was used.} The drought time course for the euphorbia, for the sentence on acid accumulation under drought.}
\rimagebox{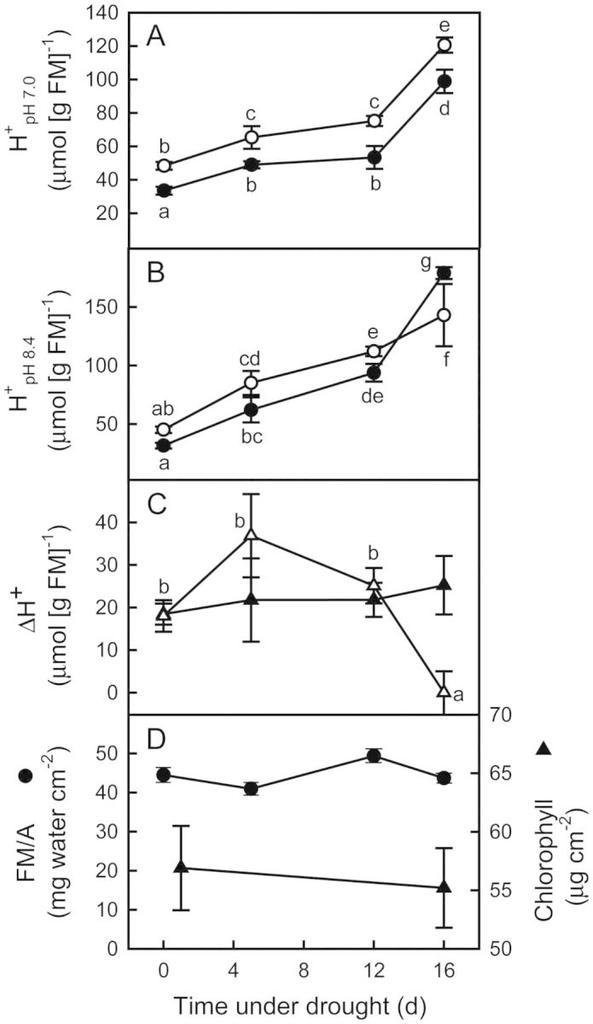}{Time course of changes with drought in E. milii leaves showing H\textsuperscript{+} content at pH 7.0 and 8.4, nocturnal H\textsuperscript{+} accumulation, and chlorophyll content}{\rbox{0.08}{0.27}{0.88}{0.49}{}}
Under drought, \emph{E. milii} shows increased nocturnal H\textsuperscript{+} accumulation, but the pattern differs from strong CAM, with citrate contributing significantly to acidification \rcite{5}.\rimgref{12}

\rsec{Phylogenetic Distribution and Evolutionary Origins}
The phylogenetic distribution of CAM reveals its polyphyletic nature. In the Portulacineae plus Molluginaceae, CAM has evolved multiple times independently \rcite{2}.

\rtrace{turn 5 \texttt{open(url)} \(\rightarrow\) turn 6 \texttt{open(id=page\_img\_117)}}{Source page: \emph{The diverse diaspora of CAM: a pole-to-pole sketch}; caption on the page: ``Presence of C3, C3 + CAM, strong CAM, C4 + CAM, C4 and C4 intermediate photosynthesis (indicated by green ticks) superimposed upon phylogenies of the Portulacineae plus Molluginaceae (upper panel) and the Montiaceae (lower panel).''\par\textbf{Why it was used.} Places the photosynthetic types on a phylogeny that includes the Cactaceae, for the section on how often CAM arose.}
\rimagebox{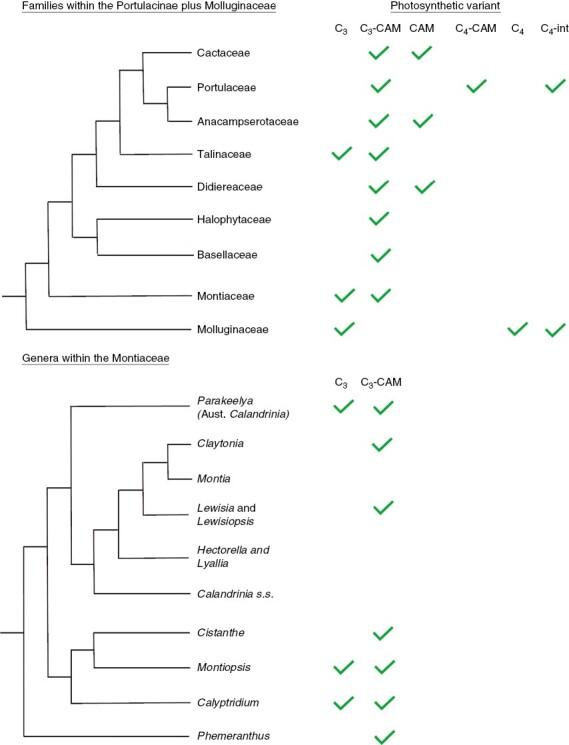}{Phylogeny of Portulacineae plus Molluginaceae showing distribution of C3, C3+CAM, CAM, C4+CAM, C4, and C4-intermediate photosynthesis}{}
Within the Montiaceae, \fact{different genera show different photosynthetic variants, with some species exhibiting C3+CAM while others are strictly C3} \rcite{2}.\rimgref{13} This pattern suggests that CAM can evolve within lineages that are otherwise C3, and that intermediate forms like C3+CAM are common stepping stones.

\rsec{Conclusions}
The evidence from anatomical cross-sections, carbon isotope ratios, and nocturnal gas exchange demonstrates that CAM photosynthesis represents a remarkable case of convergent evolution. Unrelated desert plant lineages---agaves, cacti, and euphorbias---have independently evolved similar solutions to aridity: nocturnal CO\textsubscript{2} fixation, succulent tissue for malate storage, and temporal separation of carbon acquisition from fixation.

Anatomical studies reveal convergent leaf structures with large vacuoles for acid storage and modified mesophyll organization. Carbon isotope ratios provide quantitative measures of CAM contribution, revealing a continuum from C3 through C3+CAM to strong CAM. Nocturnal gas exchange measurements confirm the characteristic CAM rhythm and demonstrate the metabolic costs of this pathway, including elevated nocturnal respiration.

The diversity of CAM expression---from obligate CAM in agaves to facultative CAM in ferns and CAM-cycling in \emph{Euphorbia}---demonstrates that this pathway can be integrated into many physiological and ecological contexts. As climate change increases water stress globally, understanding the convergent evolution of CAM may inform efforts to engineer drought tolerance into crops.

\rsec{References}
\rcite{1} Crassulacean acid metabolism (CAM) at the crossroads: a special issue to honour 50 years of CAM research by Klaus Winter\rurl{https://pmc.ncbi.nlm.nih.gov/articles/PMC10799977/}

\rcite{2} The diverse diaspora of CAM: a pole-to-pole sketch\rurl{https://pmc.ncbi.nlm.nih.gov/articles/PMC10800000/}

\rcite{3} Prospects and perspectives: inferring physiological and regulatory targets for CAM from molecular and modelling approaches\rurl{https://pmc.ncbi.nlm.nih.gov/articles/PMC10799989/}

\rcite{4} The Agavoideae: an emergent model clade for CAM evolutionary biology\rurl{https://pmc.ncbi.nlm.nih.gov/articles/PMC10799990/}

\rcite{5} Crassulacean acid metabolism-cycling in Euphorbia milii\rurl{https://pmc.ncbi.nlm.nih.gov/articles/PMC3628315/}

\rcite{6} How Closely Do the \(\delta\)\textsuperscript{1}\textsuperscript{3}C Values of Crassulacean Acid Metabolism Plants Reflect the Proportion of CO\textsubscript{2} Fixed during Day and Night?\rurl{https://pmc.ncbi.nlm.nih.gov/articles/PMC166772/}

\rcite{7} Nocturnal versus diurnal CO\textsubscript{2} uptake: how flexible is Agave angustifolia?\rurl{https://pmc.ncbi.nlm.nih.gov/articles/PMC4085958/}

\rcite{8} Gas exchange and leaf anatomy of a C3--CAM hybrid, Yucca gloriosa (Asparagaceae)\rurl{https://pmc.ncbi.nlm.nih.gov/articles/PMC4762382/}

\rcite{9} Elevated nocturnal respiratory rates in the mitochondria of CAM plants: current knowledge and unanswered questions\rurl{https://pmc.ncbi.nlm.nih.gov/articles/PMC10799998/}

\rimgsources{\rimgsrc{Image~1}{Prospects and perspectives: inferring physiological and regulatory targets for CAM from molecular and modelling approaches}{https://pmc.ncbi.nlm.nih.gov/articles/PMC10799989/}
\rimgsrc{Image~2}{The Agavoideae: an emergent model clade for CAM evolutionary biology}{https://pmc.ncbi.nlm.nih.gov/articles/PMC10799990/}
\rimgsrc{Image~3}{Gas exchange and leaf anatomy of a C3--CAM hybrid, Yucca gloriosa (Asparagaceae)}{https://pmc.ncbi.nlm.nih.gov/articles/PMC4762382/}
\rimgsrc{Images~4, 12}{Crassulacean acid metabolism-cycling in Euphorbia milii}{https://pmc.ncbi.nlm.nih.gov/articles/PMC3628315/}
\rimgsrc{Images~5, 6}{How Closely Do the \(\delta^{13}\)C Values of Crassulacean Acid Metabolism Plants Reflect the Proportion of CO\textsubscript{2} Fixed during Day and Night?}{https://pmc.ncbi.nlm.nih.gov/articles/PMC166772/}
\rimgsrc{Images~7, 11, 13}{The diverse diaspora of CAM: a pole-to-pole sketch}{https://pmc.ncbi.nlm.nih.gov/articles/PMC10800000/}
\rimgsrc{Image~8}{Nocturnal versus diurnal CO\textsubscript{2} uptake: how flexible is Agave angustifolia?}{https://pmc.ncbi.nlm.nih.gov/articles/PMC4085958/}
\rimgsrc{Images~9, 10}{Elevated nocturnal respiratory rates in the mitochondria of CAM plants: current knowledge and unanswered questions}{https://pmc.ncbi.nlm.nih.gov/articles/PMC10799998/}}
\rgate{\item Submission 1 (turn 10): {\color{tok}\textbf{OK}} --- 7,377 chars, 13 images, 9 sources}

\end{reportcard}

\end{document}